\documentclass{article}

 \usepackage[preprint]{neurips_2025}

\usepackage[utf8]{inputenc} % allow utf-8 input
\usepackage[T1]{fontenc}    % use 8-bit T1 fonts
\usepackage{url}            % simple URL typesetting
\usepackage{booktabs}       % professional-quality tables
\usepackage{hyperref}
\usepackage{amsfonts}       % blackboard math symbols
\usepackage{nicefrac}       % compact symbols for 1/2, etc.
\usepackage{xcolor}         % colors
\usepackage{amsmath}
\usepackage{algorithm}
\usepackage{algpseudocode}
\usepackage{graphicx} % for \resizebox
\usepackage{multirow} % for multirow cells
\usepackage{lipsum} % for dummy text
\usepackage{enumitem}
\usepackage{amsthm}
\usepackage{caption}
\usepackage{tabularx}
\usepackage{soul}
\usepackage[normalem]{ulem}
\usepackage{booktabs}

\newtheorem{theorem}{Theorem}[section]
\newtheorem{assumption}[theorem]{Assumption}

\title{SURE Guided Posterior Sampling: Trajectory Correction for Diffusion-Based Inverse Problems}

\author{%
  Minwoo Kim, Hongki Lim\thanks{Corresponding author.}\\
   Department of Electrical and Computer Engineering\\
  Inha University\\
  Incheon 22212, South Korea\\
  \texttt{ququlza1520@inha.edu, hklim@inha.ac.kr} \\
}

\begin{document}

\maketitle
\begin{abstract}
  Diffusion models have emerged as powerful learned priors for solving inverse problems. However, current iterative solving approaches—which alternate between diffusion sampling and data consistency steps—typically require hundreds or thousands of steps to achieve high-quality reconstruction due to accumulated errors. We address this challenge with SURE Guided Posterior Sampling (SGPS), a method that corrects sampling trajectory deviations using Stein's Unbiased Risk Estimate (SURE) gradient updates and PCA-based noise estimation. By mitigating noise-induced errors during the critical early and middle sampling stages, SGPS enables more accurate posterior sampling and reduces error accumulation. This allows our method to maintain high reconstruction quality with fewer than 100 Neural Function Evaluations (NFEs). Our extensive evaluation across diverse inverse problems demonstrates that SGPS consistently outperforms existing methods at low NFE counts.
\end{abstract}
\section{Introduction}
\label{sec:intro}

Recent advances in generative models have revolutionized computer vision, enabling high-quality data synthesis across domains such as 3D reconstruction~\citep{poole2022dreamfusion, liu2023zero1to3, anciukevivcius2023renderdiffusion}, text-to-4D~\cite{ling2024alignyourgaussians}, video generation~\cite{ho2022video, he2023latentvideodiffusionmodels, singermake}, and image generation~\cite{rombach2021highresolution, Lipman2022FlowMF, zhen2025token}. These developments have fueled interest in using diffusion models for conditional generation within an unsupervised learning framework, where data is generated based on specific constraints.

Unsupervised conditional generation offers key advantages over supervised methods by eliminating task-specific training, reducing costs, and improving generalizability. Inverse problems, a core category of such tasks, aim to recover an unknown signal $\mathbf{x}$ from a measurement $\mathbf{y}$, modeled as $\mathbf{y} = \mathbf{A}\mathbf{x} + \mathbf{n}$, where $\mathbf{A}$ is the forward operator and $\mathbf{n}$ is noise. Solving these unsupervised using diffusion models—termed Diffusion-based Inverse Problem Solvers (DIS)—involves sampling from the posterior $p(\mathbf{x}|\mathbf{y})$ using the prior $p(\mathbf{x})$ learned by the model.

% DIS methods fall into three categories~\cite{Daras2024ASO}: (1) Explicit Approximations for Measurement Matching~\cite{songscore, chungdiffusion, song2023pseudoinverse, DDRM, DDNM, DDS}, which approximate measurement consistency; (2) Variational Inference~\cite{mardanivariational, Feng2023ScoreBasedDM}, optimizing the posterior via variational techniques; and (3) Asymptotically Exact Methods~\cite{pnpdm, Dou2024DiffusionPS, Cardoso2023MonteCG}, using MCMC or SMC for true posterior sampling. While robust, Explicit Approximation methods suffer from accumulated errors due to (1) high noise in early diffusion stages distorting denoiser outputs and (2) conditional guidance introducing additional noise, misaligning actual noise levels from the expected schedule (see Fig.~\ref{geometry}). This error cascade often requires hundreds or thousands of Neural Function Evaluations (NFEs) for satisfactory results.

\begin{figure}
    \centering
    \includegraphics[width=0.7\linewidth]{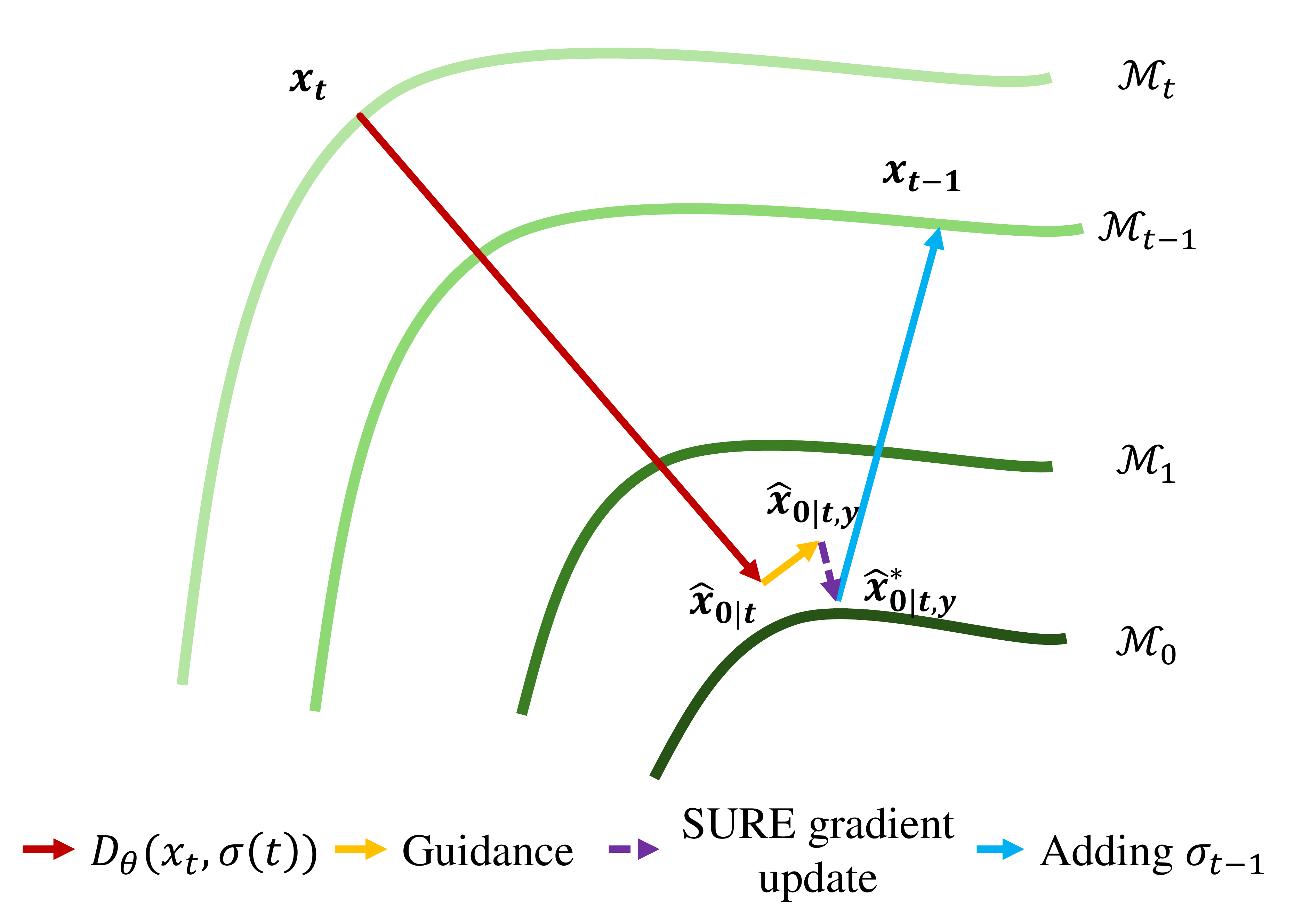}
    \caption{Geometric illustration of SGPS. Green curves represent data manifolds $\mathcal{M}_t$ at varying noise levels. Practical sampling deviates from ideal paths: diffusion denoising (red arrow) produces $\hat{\mathbf{x}}_{0|t}$ with errors from high initial noise, while conditional guidance (yellow arrow) creates further deviation by moving to $\hat{\mathbf{x}}_{0|t,\mathbf{y}}$ for measurement consistency but away from the true manifold $\mathcal{M}_0$. SGPS applies a SURE gradient update (purple arrow) to correct these deviations, yielding $\hat{\mathbf{x}}^*_{0|t,\mathbf{y}}$ closer to $\mathcal{M}_0$. Adding noise $\sigma_{t-1}$ (blue arrow) produces $\mathbf{x}_{t-1}$ with reduced accumulated error.}
    \label{geometry}
\end{figure}

DIS methods fall into four categories~\cite{Daras2024ASO}: 
(1) Explicit Approximations for Measurement Matching~\cite{songscore, song2023pseudoinverse}, which approximate measurement consistency; 
(2) Variational Inference~\cite{mardanivariational, Feng2023ScoreBasedDM, moufad2025variational}, optimizing the posterior via variational techniques; 
(3) CSGM-type methods~\cite{dmplug}, which backpropagate through a deterministic diffusion sampler to optimize the initial noise for reconstruction; and 
(4) Asymptotically Exact Methods~\cite{pnpdm, Cardoso2023MonteCG, dpnp}, using MCMC or SMC for true posterior sampling. 
While robust, Explicit Approximation methods suffer from accumulated errors due to (1) high noise in early diffusion stages distorting denoiser outputs and (2) conditional guidance introducing additional noise, misaligning actual noise levels from the expected schedule (see Fig.~\ref{geometry}). 
This error cascade often requires hundreds or thousands of Neural Function Evaluations (NFEs) for satisfactory results.

Efforts like DAPS~\cite{DAPS} (decoupling sample dependencies) and MPGD~\cite{MPGD} (projecting guidance onto the manifold’s tangent space) reduce errors but still demand over 100 NFEs, limiting practicality in resource-constrained settings. We propose SURE-Guided Posterior Sampling (SGPS) to address this by correcting sampling deviations after conditional guidance. Using Stein’s Unbiased Risk Estimate (SURE)~\cite{charles_stein_1981} to estimate MSE between true and reconstructed data, SGPS applies gradient updates to minimize errors. PCA-based noise estimation~\cite{chen2015efficient} further quantifies deviations from the intended noise schedule, ensuring accurate corrections and aligning samples with the true data manifold.

We conducted extensive evaluations across various inverse problem tasks, demonstrating that our approach consistently delivers comparable or superior results to existing methods. Fig.~\ref{overview} provides a visual overview of our SGPS sampling process. The main contributions of our work are:
\begin{itemize}
    \item We propose the SGPS algorithm, which employs SURE gradients and PCA-based noise level estimation to reduce noise and correct sampling trajectories during posterior sampling.
    
    \item We theoretically validate SGPS trajectory correction, showing conditional guidance preserves SURE-compatible noise properties and SURE-guided updates provably reduce KL divergence to the target posterior.
    
    \item We demonstrate empirically that SGPS outperforms existing methods on general inverse problems, even with fewer than 100 NFEs. Our validation of PCA-based noise level estimation shows that noise introduced during posterior sampling increases noise levels beyond the intended schedule.
\end{itemize}

The paper is structured as follows: Section~\ref{sec:background} covers diffusion models, inverse problems, and statistical foundations; Section~\ref{sec:method} details SGPS; Section~\ref{sec:experiment} presents results; and Section~\ref{sec:conclusion} discusses implications and future work.

\section{Background}
\label{sec:background}

This section reviews key concepts for understanding our approach: diffusion models for inverse problems, DAPS, Stein's Unbiased Risk Estimate, and PCA-based noise level estimation.

\subsection{Diffusion Models and Inverse Problems}

Diffusion models\cite{Song2020ImprovedTF,Ho2020DenoisingDP,Song2020DenoisingDI} reverse a gradual noising process through forward and backward processes. The forward process is formalized as:
\begin{equation}
d\mathbf{x}_t = \sqrt{2\dot{\sigma}(t)\sigma(t)} \, d\mathbf{w}_t
\end{equation}
Here, $\mathbf{x}_t$ is the noisy data at time $t$, $\dot{\sigma}(t)$ is the time derivative of the noise schedule, and $\mathbf{w}_t$ is the standard Wiener process. In this formulation, the drift term is zero, and the diffusion is controlled entirely by the noise schedule $\sigma(t)$.

The backward process for sampling is:
\begin{equation}
d\mathbf{x}_t = -2\dot{\sigma}(t)\sigma(t) \nabla_{\mathbf{x}_t} \log p_t(\mathbf{x}_t; \sigma_t) \, dt + \sqrt{2\dot{\sigma}(t)\sigma(t)} \, d\mathbf{w}_t
\end{equation}
where $\nabla_{\mathbf{x}_t} \log p_t(\mathbf{x}_t)$ is the score function guiding denoising.

For inverse problems (recovering $\mathbf{x} \in \mathbb{R}^n$ from $\mathbf{y} \in \mathbb{R}^m$ via $\mathbf{y} = \mathbf{A} \mathbf{x} + \mathbf{n}$), the reverse SDE is modified to sample from $p(\mathbf{x} \mid \mathbf{y})$:
\begin{equation}
d\mathbf{x}_t = -2\dot{\sigma}(t)\sigma(t) \nabla_{\mathbf{x}_t} \log p_t(\mathbf{x}_t \mid \mathbf{y}) \, dt + \sqrt{2\dot{\sigma}(t)\sigma(t)} \, d\mathbf{w}_t
\end{equation}

The conditional score is decomposed via Bayes' rule:
\begin{equation}
\nabla_{\mathbf{x}_t} \log p_t(\mathbf{x}_t \mid \mathbf{y}) = \nabla_{\mathbf{x}_t} \log p_t(\mathbf{x}_t) + \nabla_{\mathbf{x}_t} \log p_t(\mathbf{y} \mid \mathbf{x}_t)
\end{equation}
where the first term, $\nabla_{\mathbf{x}_t} \log p_t(\mathbf{x}_t)$, is directly provided by the pre-trained diffusion model, while the second term, $\nabla_{\mathbf{x}_t} \log p_t(\mathbf{y} \mid \mathbf{x}_t)$, is intractable because the measurement $\mathbf{y}$ relates to the clean signal $\mathbf{x}_0$ rather than the noisy intermediate sample $\mathbf{x}_t$. Different methods address this challenge through various approximation techniques~\citep{DDRM,DDNM,DDS}.

\subsection{EDM}

The Elucidating Diffusion Models (EDM) framework~\citep{Karras2022ElucidatingTD} parameterizes the diffusion process with noise level $\sigma$, simplifying the trajectory from $\sigma_{\text{min}}\approx 0$ (clean data) to $\sigma_{\text{max}}$ (pure noise). This direct noise-level parameterization is convenient for our method as it allows direct use of our estimated noise levels.

\subsection{DAPS}

DAPS~\cite{DAPS} addresses the inability to correct global errors in early diffusion steps by decoupling consecutive samples in the trajectory. For each noise level, DAPS operates in three phases: (1) estimating $\hat{\mathbf{x}}^{\mathrm{ode}}_{0|t}$ by solving a probability flow ODE with the diffusion model, (2) sampling $\hat{\mathbf{x}}_{0|t,\mathbf{y}} \sim p(\mathbf{x}_0 | \mathbf{x}_t, \mathbf{y})$ via Langevin dynamics to enforce measurement consistency, and (3) adding noise to obtain $\mathbf{x}_{t-1} \sim \mathcal{N}(\hat{\mathbf{x}}_{0|t,\mathbf{y}}, \sigma^2_{t-1}\mathbf{I})$. While DAPS improves performance on nonlinear inverse problems, it still requires $>$100 NFEs for high-quality results.

\subsection{Stein's Unbiased Risk Estimate (SURE)}

SURE~\cite{charles_stein_1981, Edupuganti2019UncertaintyQI} provides a way to estimate the Mean Squared Error (MSE) between a denoised image and its unknown ground truth without requiring access to the ground truth data. This property makes it particularly valuable for our application.

Given a ground truth image \( \mathbf{x}_0 \) and a noisy observation \( \mathbf{x}_{\text{noisy}} = \mathbf{x}_0 + \mathbf{z} \) where \( \mathbf{z} \sim \mathcal{N}(0, \sigma^2 \mathbf{I}) \), SURE is formulated as:
\begin{equation}
\text{SURE} = -n \sigma^2 + \| \hat{\mathbf{x}} - \mathbf{x}_{\text{noisy}} \|^2 + 2\sigma^2 \, \text{tr} \left( \frac{\partial \hat{\mathbf{x}}}{\partial \mathbf{x}_{\text{noisy}}} \right)
\end{equation}
where \( \hat{\mathbf{x}} = f(\mathbf{x}_{\text{noisy}}) \) is the output of a denoiser function \( f \), \( n \) is the dimension of the image, and \( \sigma^2 \) is the noise variance. A key assumption in SURE is that the noise follows a Gaussian distribution—an assumption that naturally aligns with diffusion models, which also employ Gaussian noise in their forward process, making SURE particularly suitable for correcting deviations in diffusion sampling trajectories.

Computing the Jacobian trace term directly is computationally intensive, especially for complex neural network denoisers. Therefore, an efficient Monte Carlo approximation is typically employed~\cite{Metzler2018UnsupervisedLW,Ramani2008MonteCarloSA}:
\begin{equation}
\text{tr}\{ J \} \approx \mathbf{b}^T \left( f(\mathbf{x}_{\text{noisy}} + \epsilon \mathbf{b}) - f(\mathbf{x}_{\text{noisy}}) \right) \epsilon^{-1}
\end{equation}
where \( \mathbf{b} \) is sampled from \( \mathcal{N}(0, 1) \) and \( \epsilon \) is carefully chosen hyperparameter (typically the maximum pixel value of \( \mathbf{x}_{\text{noisy}} \) divided by 1000). The derivation and proofs related to SURE are provided in the Appendix~\ref{appendix:sure_derivation}.

\subsection{PCA-based Noise Level Estimation}
Accurate noise estimation is essential for effective SURE application. We adopt a patch-based PCA approach~\cite{chen2015efficient}, which estimates noise variance by analyzing eigenvalues associated with redundant image dimensions. The method works by extracting image patches, computing their covariance matrix, and performing eigenvalue decomposition. The noise level is determined by iteratively excluding the largest eigenvalues until the mean equals the median of the remaining eigenvalues. This approach provides a computationally efficient and theoretically justified estimation without requiring additional training, making it ideal for our diffusion-based framework.
\begin{figure}
    \centering
    \includegraphics[width=\textwidth]{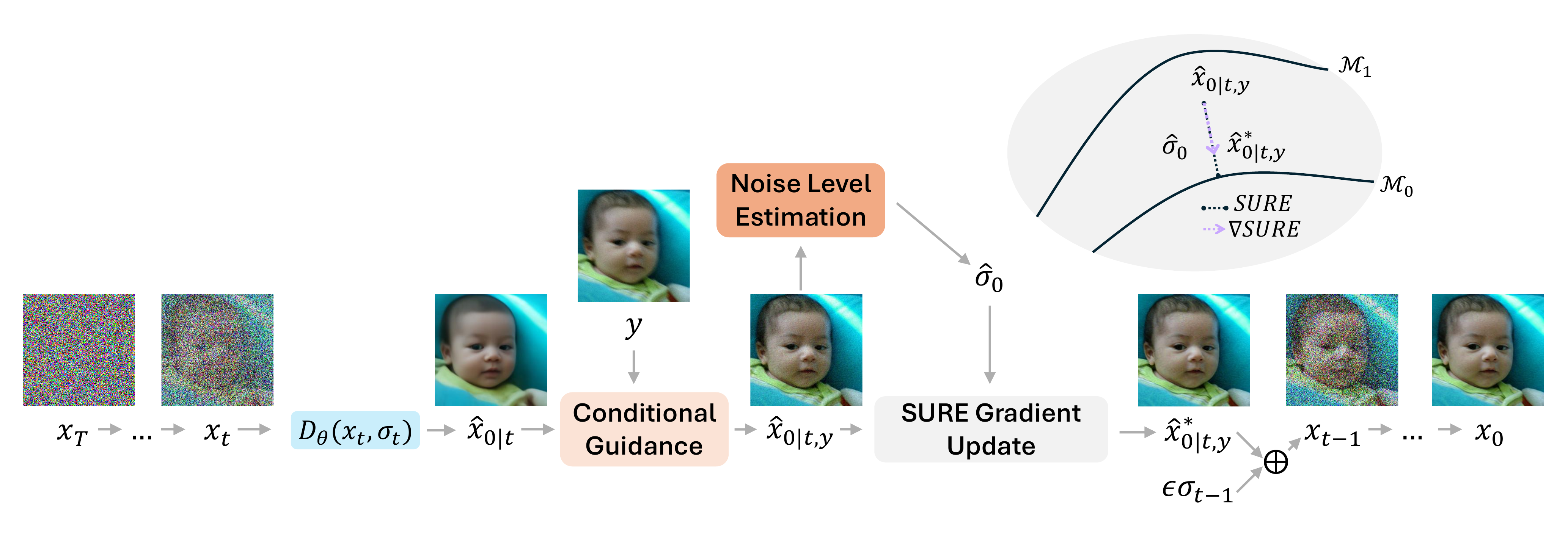}
    \caption{Overview of the SGPS sampling process. Starting with noisy sample $\mathbf{x}_t$, the process applies: (1) diffusion denoising to produce $\hat{\mathbf{x}}_{0|t}$, (2) conditional guidance using measurement $\mathbf{y}$ to yield $\hat{\mathbf{x}}_{0|t,\mathbf{y}}$, (3) PCA-based noise level estimation to determine $\hat{\sigma}_0$, (4) SURE gradient update to correct the trajectory to $\hat{\mathbf{x}}^*_{0|t,\mathbf{y}}$, and (5) addition of noise level $\sigma_{t-1}$ to obtain $\mathbf{x}_{t-1}$ for the next step. Image examples show the evolution from noise to clean reconstruction.}
    \label{overview}
\end{figure}
\section{Method}
\label{sec:method}
In theory, the diffusion denoiser's output $\hat{\mathbf{x}}_{0|t}$ should have zero noise level; however, practical constraints result in trained diffusion models introducing residual noise. Additionally, noise introduced during conditional guidance causes samples to deviate from the true data manifold. This deviation, empirically validated in Fig.~\ref{noise_influx}, complicates accurate estimation of the posterior sampling trajectory, even though subsequent diffusion steps attempt to realign samples with the true manifold.
 
Our approach addresses the root cause of error accumulation in diffusion sampling. We use SURE to measure the discrepancy between the sample obtained after conditional guidance and the true data manifold, then adjust the reconstructed sample to minimize this discrepancy, effectively aligning it closer to the theoretical noise level.
 
\subsection{SGPS Algorithm Components}
 
Our algorithm consists of four key components, each addressing specific challenges in diffusion-based inverse problem solving. As illustrated in Fig.~\ref{geometry} and Fig.~\ref{overview}, SGPS maintains closer alignment with the true data manifold throughout the sampling process, unlike previous methods that allow deviations to accumulate.
 
\paragraph{Denoising with Diffusion Denoiser}
Building on the diffusion models described in Section~\ref{sec:background}, we utilize the Elucidating Diffusion Models (EDM) framework~\cite{Karras2022ElucidatingTD}. The probability flow ODE for an SDE that induces a probability path of the marginal distribution $\mathbf{x}_t$ is represented as:
\begin{equation}
d\mathbf{x} = -\dot{\sigma}(t) \, \sigma(t) \, \nabla_{\mathbf{x}} \log p(\mathbf{x}; \sigma(t)) \, dt
\end{equation}
EDM selects $\sigma(t)=t$, making the framework more intuitive. The denoiser function $D(\mathbf{x}; \sigma)$ is estimated by a neural network $D_\theta(\mathbf{x}; \sigma)$ trained to minimize the expected $L_2$ denoising error:
\begin{equation}
\mathbb{E}_{\mathbf{x} \sim p_{\text{data}}} \mathbb{E}_{\mathbf{n} \sim \mathcal{N}(0, \sigma^2 \mathbf{I})} \left\| D(\mathbf{x} + \mathbf{n}; \sigma) - \mathbf{x} \right\|_2^2
\end{equation}
The relationship between the denoiser and score function is:
\begin{equation}
\nabla_{\mathbf{x}} \log p(\mathbf{x}; \sigma) = (D(\mathbf{x}; \sigma) - \mathbf{x}) / \sigma^2
\end{equation}
Using this relationship, we can approximate the score function and solve the corresponding ODE to estimate the clean image $\hat{\mathbf{x}}_{0|t}$ through multi-step numerical integration such as Euler's method. However, our empirical investigations show that a single-step approach, directly applying the denoiser as $\hat{\mathbf{x}}_{0|t} = D_\theta(\mathbf{x}_t, \sigma_t)$, achieves comparable reconstruction quality while significantly reducing computational cost.
 
\paragraph{Conditional Guidance}
Our method offers flexibility—any guidance approach ensuring data consistency can be utilized. We chose Langevin dynamics~\cite{Welling2011BayesianLV,Song2021SolvingIP,Song2019GenerativeMB} for computational efficiency.
 
To sample from $p(\mathbf{x}_0 | \mathbf{x}_t, \mathbf{y})$, we apply Bayes' rule:
\begin{equation}
p(\mathbf{x}_0 | \mathbf{x}_t, \mathbf{y}) \propto p(\mathbf{x}_0 | \mathbf{x}_t) \cdot p(\mathbf{y} | \mathbf{x}_0)
\end{equation}
 
Taking the gradient of the log-posterior:
\begin{equation}
\nabla_{\mathbf{x}_0} \log p(\mathbf{x}_0 | \mathbf{x}_t, \mathbf{y}) = \nabla_{\mathbf{x}_0} \log p(\mathbf{x}_0 | \mathbf{x}_t) + \nabla_{\mathbf{x}_0} \log p(\mathbf{y} | \mathbf{x}_0)
\end{equation}
 
With Gaussian approximations $p(\mathbf{x}_0 | \mathbf{x}_t) \approx \mathcal{N}(\mathbf{x}_0; \hat{\mathbf{x}}_{0|t}, \sigma_t^2 \mathbf{I})$ and $p(\mathbf{y} | \mathbf{x}_0) = \mathcal{N}(\mathbf{y}; A(\mathbf{x}_0), \sigma_y^2 \mathbf{I})$, the Langevin dynamics update becomes:
\begin{equation}
\hat{\mathbf{x}}_0^{(j+1)} = \hat{\mathbf{x}}_0^{(j)} - \eta \cdot \nabla_{\hat{\mathbf{x}}_0^{(j)}} \left( \frac{\|\hat{\mathbf{x}}_0^{(j)} - \hat{\mathbf{x}}_{0|t}\|^2}{2\sigma_t^2} + \frac{\|A(\hat{\mathbf{x}}_0^{(j)}) - \mathbf{y}\|^2}{2\sigma_y^2} \right) + \sqrt{2\eta} \mathbf{\epsilon}_j, \quad \mathbf{\epsilon}_j \sim \mathcal{N}(0, \mathbf{I})
\label{eq:langevin dynamics}
\end{equation}
 
As \(\eta \to 0\) and \(j \to \infty\), this sequence converges to a sample \(\hat{\mathbf{x}}_{0|t,\mathbf{y}}\) consistent with both \(\hat{\mathbf{x}}_{0|t}\) and the measurement \(\mathbf{y}\). In practice, a finite number of iterations $j$ suffices for an approximate solution.
 
% \begin{theorem}[Single-Step Gaussian Preservation]
% \label{thm:one-step-gaussian}
% Under the following assumptions:
% \begin{enumerate}
%   \item Locally Lipschitz Likelihood: $\|\nabla\log p(\mathbf{y}|\mathbf{u}) - \nabla\log p(\mathbf{y}|\mathbf{v})\| \leq L_\ell\|\mathbf{u}-\mathbf{v}\|$ for any $\mathbf{u},\mathbf{v}$ in a domain containing iterates with high probability.
%   \item Incoming Gaussian: $\hat{\mathbf{x}}_{0|t} \sim \mathcal{N}(\mathbf{m},\sigma^2\mathbf{I})$ (or close).
% \end{enumerate}
% Let $\eta > 0$ be small (e.g., $\eta \leq 0.5$). Define
% \[
%    R := \hat{\mathbf{x}}_{0|t,\mathbf{y}} - \mathbf{m},
%    \quad
%    \hat{\mathbf{x}}_{0|t,\mathbf{y}} := \hat{\mathbf{x}}_{0|t}
%    - \eta\,\nabla\log p(\mathbf{y}|\hat{\mathbf{x}}_{0|t})
%    + \sqrt{2\eta}\,\sigma\,\boldsymbol{\xi},
% \]
% and $G_{target} \sim \mathcal{N}(0,\sigma^2\mathbf{I})$. Then:
% \begin{enumerate}
%   \item $W_2^2(\mathcal{L}(R),\,G_{target})=O(\eta^2\,d\,\sigma^2),$
%   \item $\mathrm{TV}(\mathcal{L}(R),\,G_{target})=O(\eta),$
%   \item $\chi^2(\mathcal{L}(R)\|\;G_{target})=O(\eta^2)$.
% \end{enumerate}
% \end{theorem}

\begin{theorem}[Gaussian Preservation in Diffusion Sampling]
\label{thm:gaussian-preservation}
Under the following assumptions:
\begin{enumerate}
  \item Locally Lipschitz Likelihood: $\|\nabla\log p(\mathbf{y}|\mathbf{u}) - \nabla\log p(\mathbf{y}|\mathbf{v})\| \leq L_\ell\|\mathbf{u}-\mathbf{v}\|$ 
  \item Incoming Gaussian: $\hat{\mathbf{x}}_{0|t} \sim \mathcal{N}(\mathbf{m}_t, \sigma_{\mathrm{residual}}^2\mathbf{I})$ (or close) ,where $\mathbf{m}_t := \mathbb{E}[\hat{\mathbf{x}}_{0|t}]$
  \item Small step size (e.g., $\eta \leq 0.5$) in each Langevin step
\end{enumerate}
 
For a single step from $t$ to $t-1$:
Let $\hat{\mathbf{x}}_{0|t,\mathbf{y}} := \hat{\mathbf{x}}_{0|t} + \eta\,\nabla\log p(\mathbf{y}|\hat{\mathbf{x}}_{0|t}) + \sqrt{2\eta}\,\sigma_t\,\boldsymbol{\xi}$, and $R_t := \hat{\mathbf{x}}_{0|t,\mathbf{y}} - \mathbf{m}_t$. Here $\mathcal{L}(R_t)$ denotes the probability distribution of $R_t$. Then $R_t$ remains close to $G_{target} \sim \mathcal{N}(0,\sigma_t^2\mathbf{I})$ with Wasserstein-2 distance bounded by:
\begin{equation}
W_2^2(\mathcal{L}(R_t),\,G_{target})=O(\eta^2\,n\,\sigma_t^2)
\end{equation}
 
For the multi-step case with $K$ steps:
Let $\mathbf{x}_{t-k} \sim \mathcal{N}(\hat{\mathbf{x}}^*_{0|t-k+1,\mathbf{y}}, \sigma_{t-k}^2\mathbf{I})$ for $k=1,2,...,K$, where $\hat{\mathbf{x}}^*_{0|t-k+1,\mathbf{y}}$ is the SGPS-corrected estimate. Then for any $k \leq K$, the deviation of $\hat{\mathbf{x}}_{0|t-k,\mathbf{y}}$ from its mean remains approximately Gaussian, with bounded divergence from the Gaussian distribution that depends on the number of steps and the step size:
\begin{equation}
W_2^2(\mathcal{L}(\hat{\mathbf{x}}_{0|t-k,\mathbf{y}} - \mathbb{E}[\hat{\mathbf{x}}_{0|t-k,\mathbf{y}}]),\,\mathcal{N}(0,\sigma_{t-k}^2\mathbf{I})) = O(K\eta^2\,d\,\max_{i\leq K}\sigma_{t-i}^2)
\end{equation}
\end{theorem}
 
The assumption that $\hat{\mathbf{x}}_{0|t} \sim \mathcal{N}(\mathbf{m}_t, \sigma_{\mathrm{residual}}^2\mathbf{I})$ is a practical and empirically supported modeling assumption. While a theoretically optimal denoiser would perfectly predict the conditional expectation $\mathbb{E}[\mathbf{x}_0 | \mathbf{x}_t]$, practical implementations such as neural networks are trained to approximate this function and inevitably introduce their own estimation errors. These errors can be modeled as approximately Gaussian, particularly at higher noise levels where central limit theorem suggests aggregated error sources trend toward Gaussian behavior. Our own analysis (Fig.~\ref{histogram} in Appendix) confirms that residual noise in denoiser outputs exhibits near-Gaussian characteristics across diffusion steps, with the approximation becoming increasingly accurate at moderate to high noise levels—precisely the regime where error accumulation is most problematic in existing methods.
 
This theorem confirms that a small-step Langevin update as used in our conditional guidance preserves the near-Gaussian structure of the noise. This provides a theoretical foundation for the subsequent application of SURE, which assumes Gaussian noise.
 
\paragraph{Noise Level Estimation}
Noise level estimation is a critical component of our method, as the diffusion denoiser $D_\theta(\mathbf{x}, \sigma)$ is explicitly trained with time step (or noise level) information. While the diffusion model operates according to a predefined noise schedule $\sigma(t)$, the actual noise level in $\hat{\mathbf{x}}_{0|t, \mathbf{y}}$ after conditional guidance is unknown and deviates from this schedule. We denote this unknown true noise level as $\sigma_0$ and aim to estimate it accurately as $\hat{\sigma}_0$.
 
Without accurate estimation of $\hat{\sigma}_0$, the diffusion denoiser cannot be properly applied, leading to inaccurate SURE calculations and suboptimal correction. We employ the patch-based PCA method~\cite{chen2015efficient} described in Section~\ref{sec:background} to estimate this crucial parameter. This method specifically addresses the Gaussian noise characteristics present in diffusion processes, providing theoretical guarantees for both efficiency and accuracy. The estimated noise level $\hat{\sigma}_0$ is used to condition the denoiser $D_\theta(\mathbf{x}_{\text{noisy}}, \hat{\sigma}_0)$ in our SURE calculation, ensuring the denoiser operates on the actual noise present in the sample, rather than the theoretical noise level from the predefined schedule.
 
\paragraph{SURE Gradient Update}
The SURE gradient update forms our core innovation, directly addressing error accumulation during diffusion sampling. While previous approaches propagate errors through sampling steps, our correction mechanism explicitly targets deviations introduced during conditional guidance.
 
SURE naturally complements diffusion models since both operate under Gaussian noise assumptions. A natural question is why we use the SURE gradient update approach rather than simply reapplying the denoiser with the estimated noise level $\hat{\sigma}_0$. While directly reapplying the denoiser $D_{\theta}(\hat{\mathbf{x}}_{0|t,\mathbf{y}}, \hat{\sigma}_{0})$ is computationally cheaper, our SURE gradient approach offers a more principled, step-wise refinement. Unlike standard denoisers trained for optimal expected SURE across the entire data distribution, our method treats SURE as a local cost function for a specific sample.

This allows for a SURE gradient step that explicitly moves each individual sample toward its local minimum, representing a theoretically justified error correction (Theorem~\ref{thm:multistep-sgps}). By utilizing the gradient direction rather than a fixed-strength adjustment, we achieve a principled optimality in local error correction, which justifies the additional computational cost through significant quality improvements as validated in our ablation studies.

% \st{While directly reapplying the denoiser (i.e., $\hat{\mathbf{x}}^*_{0|t,\mathbf{y}} = D_\theta(\hat{\mathbf{x}}_{0|t,\mathbf{y}}, \hat{\sigma}_0)$) would be computationally cheaper, the SURE gradient approach offers several advantages: (1) it provides theoretically optimal error minimization through an unbiased estimate of the MSE to the unknown ground truth, (2) it allows for fine-tuned corrections via the step size parameter $\alpha$ rather than fixed-strength adjustments, and (3) it potentially better preserves important image details by applying corrections based on gradient directions rather than uniform denoising. Our empirical results confirm that the modest additional computational cost (one extra NFE) is justified by the significant quality improvements across diverse inverse problems.}
 
After conditional guidance, $\hat{\mathbf{x}}_{0|t, \mathbf{y}}$ contains residual noise that deviates from the predefined schedule. We treat this sample as a noisy observation $\mathbf{x}_{\text{noisy}}$ and quantify its actual noise level $\hat{\sigma}_0$ using the patch-based PCA method. This noise estimation is crucial for applying our diffusion denoiser correctly: $\hat{\mathbf{x}} = D_\theta(\mathbf{x}_{\text{noisy}}, \hat{\sigma}_0)$, ensuring optimal performance in the subsequent SURE calculations.
 
We then compute the Jacobian trace through Monte Carlo approximation:
\begin{equation}
\operatorname{tr}\{J\} = \mathbf{b}^T ( D_\theta(\mathbf{x}_{\text{noisy}} + \epsilon \mathbf{b}, \max(\epsilon, \hat{\sigma}_0)) - \hat{\mathbf{x}} ) \epsilon^{-1}
\end{equation}
where $\mathbf{b} \sim \mathcal{N}(0, \mathbf{I})$ and $\epsilon$ is a small constant.
 
This allows us to formulate SURE:
\begin{equation}
\operatorname{SURE}(t) = -n \hat{\sigma}_0^2 + \|\mathbf{x}_{\text{noisy}} - \hat{\mathbf{x}}\|^2 + 2\hat{\sigma}_0^2 \operatorname{tr}\{J\}
\end{equation}
 
Finally, we apply a gradient update to minimize this risk estimate:
\begin{equation}
\hat{\mathbf{x}}^{*}_{0|t, \mathbf{y}} = \hat{\mathbf{x}}_{0|t, \mathbf{y}} - \alpha \nabla_{\hat{\mathbf{x}}_{0|t, \mathbf{y}}} \operatorname{SURE}(t)
\end{equation}

\begin{theorem}[KL Convergence under Biased SURE Gradients]
\label{thm:multistep-sgps}
Under the following assumptions:
\begin{enumerate}
  \item Local strong convexity and smoothness of $\Phi(\mathbf{x})=-\log p(\mathbf{x}|\mathbf{y})$: $\nabla^2\Phi(\mathbf{x}) \succeq \mu \mathbf{I}$ and $\|\nabla\Phi(\mathbf{x})-\nabla\Phi(\mathbf{z})\| \leq L\|\mathbf{x}-\mathbf{z}\|$ (Here, $\mu>0$ is the (local) strong convexity constant).
  
  \item The SURE gradient update $g_{\mathrm{corr}}(\hat{\mathbf{x}}_{0|t,\mathbf{y}}) := \alpha\nabla_{\hat{\mathbf{x}}_{0|t,\mathbf{y}}}\mathrm{SURE}(t)$ approximates $\hat{\sigma}_0^2 \nabla\Phi(\hat{\mathbf{x}}_{0|t,\mathbf{y}})$ with bounded bias and variance.
  
  \item Step size $\beta_t = \alpha \hat{\sigma}_0^2$ is sufficiently small, e.g., $\beta_t \leq \frac{1}{2L}$.
\end{enumerate}
Define:
\begin{equation}
  g(\hat{\mathbf{x}}_{0|t,\mathbf{y}}) := \frac{1}{\hat{\sigma}_0^2} g_{\mathrm{corr}}(\hat{\mathbf{x}}_{0|t,\mathbf{y}}),
  \quad
  \hat{\mathbf{x}}^{*}_{0|t,\mathbf{y}} := \hat{\mathbf{x}}_{0|t,\mathbf{y}} - \beta_t\,g(\hat{\mathbf{x}}_{0|t,\mathbf{y}}),
  \quad
  \beta_t=\alpha \hat{\sigma}_0^2.
\end{equation}
Let $q_t,q_t^*$ be the distributions of $\hat{\mathbf{x}}_{0|t,\mathbf{y}}$ before/after this update.
Then there exist constants $C>0$ and a mismatch $\Delta_t$ such that:
\begin{equation}
  D_{\mathrm{KL}}(q_t^*\,\|p(\mathbf{x}|\mathbf{y}))
  \leq
  (1-\beta_t\mu)\,D_{\mathrm{KL}}(q_t\|p(\mathbf{x}|\mathbf{y}))
  + \beta_t^2\,C
  + \Delta_t.
\end{equation}
\end{theorem}

\begin{algorithm}
\caption{SURE Guided Posterior Sampling}
\label{algorithm}
\begin{algorithmic}
\Require denoiser $D_\theta$, measurement $\mathbf{y}$, noise schedule $\sigma_t$, guidance($\cdot$), $T$, $\alpha$, $\epsilon$, $\mathbf{b}$, $n$
\State Sample $\mathbf{x}_T \sim \mathcal{N}(0, \sigma_T^2 \mathbf{I})$
\For{$t = T$ to $1$}
    \State \textbf{Denoising step:}
    \State $\hat{\mathbf{x}}_{0|t} = D_\theta(\mathbf{x}_t, \sigma_t)$
    \State \textbf{Conditional guidance step:}
    \State $\hat{\mathbf{x}}_{0|t, \mathbf{y}} \leftarrow \text{guidance}(\hat{\mathbf{x}}_{0|t})$ \Comment{Using Eq.~\ref{eq:langevin dynamics}}
    \State \textbf{Noise Level Estimation step:}
    \State Extract patches $\{ \mathbf{x}_i \}_{i=1}^s$ from $\hat{\mathbf{x}}_{0|t, \mathbf{y}}$
    \State $\mu = \frac{1}{s} \sum_{i=1}^s \mathbf{x}_i$
    \vspace{0.1cm}
    \State $\Sigma = \frac{1}{s} \sum_{i=1}^s (\mathbf{x}_i - \mu)(\mathbf{x}_i - \mu)^T$
    \State Calculate eigenvalues $\lambda_1 \geq \lambda_2 \geq \dots \geq \lambda_r$ of $\Sigma$
    \For{$i = 1$ to $r$}
        \State $\tau = \frac{1}{r - i + 1} \sum_{j=i}^r \lambda_j$
        \If{$\tau$ is the median of the set $\{ \lambda_j \}_{j=i}^r$}
            \State $\hat{\sigma}_0 = \sqrt{\tau}$ and \textbf{break}
        \EndIf
    \EndFor
    \State \textbf{SURE gradient update step:}
    \State $\mathbf{x}_{\text{noisy}} = \hat{\mathbf{x}}_{0|t, \mathbf{y}}$
    \State $\hat{\mathbf{x}} = D_\theta(\mathbf{x}_{\text{noisy}}, \hat{\sigma}_0)$
    \State $\operatorname{tr}\{J\} = \mathbf{b}^T \left( D_\theta(\mathbf{x}_{\text{noisy}} + \epsilon \mathbf{b}, \max(\epsilon, \hat{\sigma}_0)) - \hat{\mathbf{x}} \right) \epsilon^{-1}$
    \State $\operatorname{SURE}(t) = -n \hat{\sigma}_0^2 + \|\mathbf{x}_{\text{noisy}} - \hat{\mathbf{x}}\|^2 + 2\hat{\sigma}_0^2 \operatorname{tr}\{J\}$
    \State $\hat{\mathbf{x}}^{*}_{0|t, \mathbf{y}} = \hat{\mathbf{x}}_{0|t, \mathbf{y}} - \alpha \nabla_{\hat{\mathbf{x}}_{0|t, \mathbf{y}}} \operatorname{SURE}(t)$
    \State Sample $\mathbf{x}_{t-1} \sim \mathcal{N}(\hat{\mathbf{x}}^{*}_{0|t, \mathbf{y}}, \sigma_{t-1}^2 \mathbf{I})$
\EndFor
\State \Return $\mathbf{x}_0$ 
\end{algorithmic}
\end{algorithm}
 
This theorem provides a theoretical guarantee that each SURE gradient update step reduces the KL divergence between the sample distribution and the true posterior distribution, confirming that our approach progressively improves sampling accuracy.

This gradient update is critical for addressing accumulated errors in the sampling trajectory. When the diffusion denoiser and conditional guidance introduce errors that cause deviation from the ideal trajectory, this update effectively pulls the sample back toward the true data manifold. By minimizing SURE—which provides an unbiased estimate of the MSE between the sample and its unknown ground truth—we correct the accumulated noise without requiring access to the ground truth data.

This correction serves two crucial purposes. First, it reduces the immediate error in the current sample, preventing the noise level from diverging from the predefined schedule. Second, it prevents error propagation to subsequent steps, addressing the cascading error accumulation problem. Without this correction, errors would compound across iterations, requiring many more steps for convergence. With SGPS, each step builds upon a more accurately placed sample, enabling high-quality reconstruction with significantly fewer NFEs. The complete SGPS algorithm is presented in Algorithm~\ref{algorithm},which integrates all these components into a coherent framework.

\section{Experiments}\label{sec:experiment}

\subsection{Experimental setup}
We largely follow the experimental setup in~\cite{DAPS}. As previously mentioned, our method produces reliable results even with fewer than 100 NFEs. For baseline methods, we run experiments using exactly 50 and 100 NFEs for all linear and nonlinear tasks. However, SGPS requires 3 NFEs per diffusion sampling step, as it evaluates the denoiser three times: once for estimating \(\hat{\mathbf{x}}_{0 \mid t}\) and twice for computing SURE. Therefore, we adjust its evaluation to 48 NFEs (\(T=16\)) and 99 NFEs (\(T=33\)), respectively, which are the closest feasible values aligning with the baseline computational budget. Additionally, since DPS showed poor performance at lower NFEs, we included its results at 1000 NFEs, following the original settings in~\cite{chungdiffusion}. In the conditional guidance step, we use 100 Langevin dynamics steps. All inference experiments were conducted on a single NVIDIA RTX 4090.

\begin{figure}
    \centering
    \includegraphics[width=\textwidth]{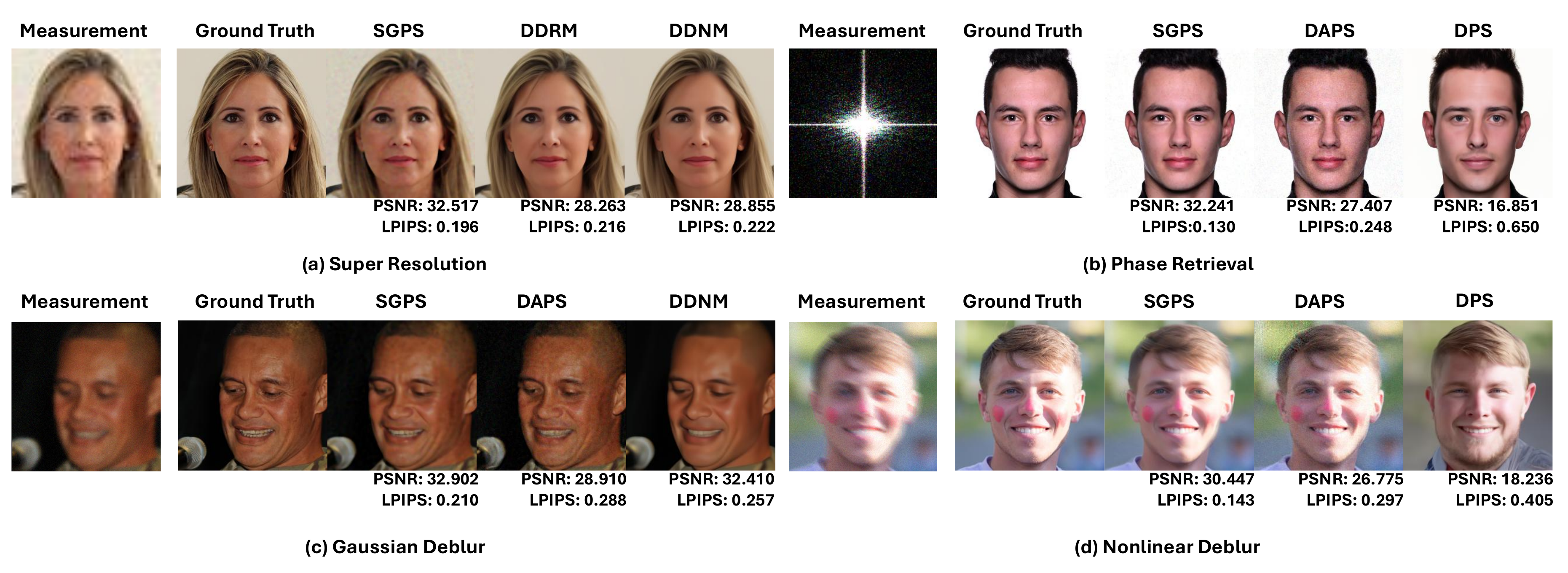}
    \caption{Qualitative results of SGPS and baseline methods on different general inverse problems. The figure compares the top-3 performing methods for each task, as reported in the NFE 50 comparisons in Table~\ref{linear_combined},~\ref{nonlinear_combined}.}
    \label{sample}
\end{figure}

\begin{table*}[ht]
    \centering
    \setlength{\tabcolsep}{3pt}
    \footnotesize
    \resizebox{0.9\textwidth}{!}{
        \begin{tabular}{l c cc cc cc cc cc}
            \toprule
            \multirow{2}{*}{Method} & \multirow{2}{*}{NFE} &
            \multicolumn{2}{c}{SR4} &
            \multicolumn{2}{c}{Inpaint(Box)} &
            \multicolumn{2}{c}{Inpaint(Random)} &
            \multicolumn{2}{c}{Gaussian deblur} &
            \multicolumn{2}{c}{Motion deblur} \\
            \cmidrule(lr){3-4}\cmidrule(lr){5-6}\cmidrule(lr){7-8}\cmidrule(lr){9-10}\cmidrule(lr){11-12}
            & & LPIPS$\downarrow$ & PSNR$\uparrow$ & LPIPS$\downarrow$ & PSNR$\uparrow$ &
                LPIPS$\downarrow$ & PSNR$\uparrow$ & LPIPS$\downarrow$ & PSNR$\uparrow$ &
                LPIPS$\downarrow$ & PSNR$\uparrow$ \\
            \midrule

            \multicolumn{12}{c}{NFE $\approx$ 100} \\
            \midrule
            SGPS (ours) & 99 & \textbf{0.179} & \textbf{29.384} & \textbf{0.133} & \underline{24.231} & \textbf{0.116} & \textbf{30.466} & \textbf{0.179} & \textbf{29.353} & \textbf{0.148} & \textbf{31.239} \\
            DAPS        & 100 & 0.230 & 27.693 & 0.192 & 22.513 & 0.238 & 26.636 & \underline{0.220} & 27.771 & \underline{0.167} & \underline{29.838} \\
            DPS         & 1000 & 0.301 & 23.730 & 0.205 & 23.146 & 0.220 & 28.089 & 0.257 & 25.012 & 0.260 & 24.865 \\
            DPS         & 100 & 0.494 & 16.739 & 0.377 & 19.828 & 0.401 & 19.954 & 0.390 & 20.103 & 0.392 & 19.983 \\
            FPS         & 100 & \underline{0.215} & 27.628 & 0.176 & \textbf{24.380} & \underline{0.167} & \underline{30.176} & 0.703 & 12.140 & - & - \\
            DDRM        & 100 & 0.246 & 28.326 & \underline{0.171} & 24.123 & 0.189 & 29.796 & 0.290 & 23.573 & - & - \\
            DDNM        & 100 & 0.256 & \underline{28.869} & 0.195 & 24.109 & 0.202 & 30.133 & 0.286 & \underline{28.653} & - & - \\
            \addlinespace[3pt]
            \midrule

            \multicolumn{12}{c}{NFE $\approx$ 50} \\
            \midrule
            SGPS (ours) & 48 & \textbf{0.198} & \textbf{29.077} & \textbf{0.139} & \textbf{23.569} & \textbf{0.177} & 27.602 & \textbf{0.192} & \textbf{29.080} & \textbf{0.165} & \textbf{30.79} \\
            DAPS        & 50 & 0.283 & 26.529 & 0.210 & 21.449 & 0.367 & 22.513 & \underline{0.267} & 26.733 & \underline{0.211} & \underline{28.530} \\
            DPS         & 50 & 0.550 & 14.688 & 0.484 & 16.410 & 0.479 & 17.312 & 0.480 & 17.138 & 0.479 & 17.165 \\
            FPS         & 50 & \underline{0.232} & 27.185 & 0.287 & \underline{23.390} & 0.219 & \underline{28.524} & 0.726 & 11.793 & - & - \\
            DDRM        & 50 & 0.248 & 28.177 & \underline{0.184} & 23.083 & \underline{0.213} & 28.508 & 0.284 & 25.233 & - & - \\
            DDNM        & 50 & 0.257 & \underline{28.664} & 0.196 & 23.508 & 0.227 & \textbf{28.756} & 0.281 & \underline{28.575} & - & - \\
            \bottomrule
        \end{tabular}
    }
    \caption{Quantitative evaluation (PSNR, LPIPS) of various methods at NFE 100 and 50 for solving linear inverse problems.}
    \label{linear_combined}

    \vspace{20pt}

    \small
    \begin{tabular}{l c cc cc cc}
        \toprule
        \multirow{2}{*}{Method} & \multirow{2}{*}{NFE} &
        \multicolumn{2}{c}{Phase retrieval} &
        \multicolumn{2}{c}{Nonlinear deblurring} &
        \multicolumn{2}{c}{High dynamic range} \\
        \cmidrule(lr){3-4}\cmidrule(lr){5-6}\cmidrule(lr){7-8}
        & & LPIPS$\downarrow$ & PSNR$\uparrow$ & LPIPS$\downarrow$ & PSNR$\uparrow$ & LPIPS$\downarrow$ & PSNR$\uparrow$ \\
        \midrule

        \multicolumn{8}{c}{NFE $\approx$ 100} \\
        \midrule
        SGPS (ours) & 99 & \textbf{0.268} & \textbf{24.080} & \textbf{0.197} & \textbf{27.332} & \textbf{0.179} & \textbf{24.872} \\
        DAPS        & 100 & \underline{0.402} & \underline{20.830} & \underline{0.255} & \underline{25.564} & \underline{0.199} & \underline{24.093} \\
        DPS         & 1000 & 0.483 & 17.406 & 0.370 & 20.678 & 0.319 & 20.778 \\
        DPS         & 100 & 0.513 & 14.257 & 0.423 & 18.945 & 0.604 & 11.012 \\
        \addlinespace[3pt]
        \midrule

        \multicolumn{8}{c}{NFE $\approx$ 50} \\
        \midrule
        SGPS (ours) & 48 & \textbf{0.378} & \textbf{20.951} & \textbf{0.218} & \textbf{26.116} & \textbf{0.187} & \textbf{23.918} \\
        DAPS        & 50 & 0.533 & \underline{17.342} & \underline{0.323} & \underline{24.277} & \underline{0.231} & \underline{22.870} \\
        DPS         & 50 & \underline{0.531} & 13.472 & 0.491 & 16.825 & 0.671 & 9.134 \\
        \bottomrule
    \end{tabular}
    \caption{Quantitative evaluation (PSNR, LPIPS) of various methods at NFE 100 and 50 for solving non-linear inverse problems.}
    \label{nonlinear_combined}
\end{table*}

\paragraph{Datasets and Metric}
We conducted experiments using the FFHQ \( 256 \times 256 \) dataset~\cite{Karras2018ASG}, frequently used in inverse problem research including diffusion models, and utilized 100 images from the FFHQ validation set. The pre-trained diffusion model employed was trained on FFHQ, provided by~\cite{chungdiffusion}. For quantitative comparison, we focus on two metrics: Peak Signal-to-Noise Ratio (PSNR), which measures the quality of image reconstruction, and Learned Perceptual Image Patch Similarity (LPIPS) score, which assesses perceptual similarity between the ground truth and the predicted samples.

\paragraph{Inverse Problems}
We evaluated SGPS on five linear inverse problems: (i) super-resolution with 4$\times$ bicubic down-sampling, 
(ii) box-type inpainting with a $128 \times 128$ mask, (iii) random inpainting with 70\% pixel masking, 
(iv) Gaussian deblurring using a $61 \times 61$ kernel ($\sigma = 3.0$), and (v) motion deblurring with a 
$61 \times 61$ kernel ($\sigma = 0.5$), randomly generated per~\cite{chungdiffusion}. For nonlinear problems, we 
assessed three tasks: (i) phase retrieval with 2.0 oversampling, selecting the best of four trials 
per~\cite{chungdiffusion}, (ii) nonlinear deblurring following~\cite{Tran2021ExploreID}, and (iii) HDR recovery 
(2$\times$) from low dynamic range images. All measurements included Gaussian noise ($\sigma = 0.05$).

\paragraph{Baselines}
We compared SGPS with the following baselines: DPS~\cite{chungdiffusion}, FPS~\cite{Dou2024DiffusionPS}, DAPS~\cite{DAPS}, DDRM~\cite{DDRM}, and DDNM~\cite{DDNM}. FPS was excluded from motion deblurring comparisons as it uses anisotropic Gaussian deblurring instead. Per~\cite{chungdiffusion}, DDRM relies on SVD and is limited to tasks like Gaussian deblurring with separable kernels, but not motion deblurring due to its complex PSF. Similarly, DDNM, with its SVD-based noisy task variant, was also omitted from motion deblurring for the same reason. For a fair comparison, we used the same pre-trained model for all other methods and adjusted the sampling steps to match 50 and 100 NFEs.

\subsection{Experimental Results}
\paragraph{Quantitative and Qualitative Results}

The quantitative results in Tables~\ref{linear_combined} and~\ref{nonlinear_combined} demonstrate that SGPS consistently outperforms baselines at comparable NFEs across both linear tasks (super-resolution, deblurring, inpainting) and challenging nonlinear problems (phase retrieval, nonlinear blur). These improvements validate our hypothesis about the Gaussian nature of noise during conditional guidance. Unlike DDRM and DDNM which rely on SVD and struggle with complex forward operators, our method efficiently solves all tested problems.

Figure~\ref{sample} illustrates SGPS's high-quality reconstructions even at 50 NFE. In Super-Resolution, DDRM and DDNM produce visually smooth but detail-deficient results (e.g., exaggerated nasolabial folds and poorly reconstructed hair fringes). In Nonlinear Deblur, DAPS shows noise artifacts while DPS fails to maintain data consistency. In contrast, SGPS achieves superior data consistency while preserving fine details across all tasks, demonstrating its ability to balance fidelity and detail preservation at low NFE counts.

\begin{table}
\captionsetup{skip=10pt}
\centering
\begin{tabular}{l ccc ccc}
\toprule
\multirow{2}{*}{Method} &
\multicolumn{3}{c}{Time $\approx$ 4s} &
\multicolumn{3}{c}{Time $\approx$ 8s} \\
\cmidrule(lr){2-4}\cmidrule(lr){5-7}
& NFE & Sec/Image & PSNR & NFE & Sec/Image & PSNR \\
\midrule
SGPS  & 48  & 4.133 & \underline{29.058} & 99  & 8.458 & \textbf{29.340} \\
DAPS  & 110 & 4.257 & 27.933 & 225 & 8.659 & 28.614 \\
DPS   & 110 & 4.301 & 16.842 & 210 & 8.527 & 18.655 \\
FPS   & 180 & 4.270 & 27.852 & 360 & 8.662 & 28.077 \\
DDRM  & 190 & 4.294 & 28.367 & 385 & 8.731 & 28.384 \\
DDNM  & 190 & 4.202 & \textbf{29.094} & 380 & 8.661 & \underline{29.291} \\
\bottomrule
\end{tabular}
\caption{\textbf{Time-Equalized Comparison of Sampling Performance for SR4 Inverse Problems.} Each method is evaluated under two inference-time budgets.}
\label{tab:time}
\end{table}

Because NFE alone does not fully capture computational cost, we additionally compare methods under the same run-time budget. As shown in Table~\ref{tab:time}, SGPS maintains strong reconstruction quality at competitive runtimes: at roughly 4 seconds, it reaches PSNR 29.058 very close to the best result from DDNM (29.094) and at 8 seconds it attains PSNR 29.340, matching or exceeding the other baselines in our setting. A runtime breakdown indicates that the primary overhead comes from the SURE gradient update, accounting for about 51.2\% of the total time (22.4\% from denoiser evaluations and 30\% from autograd-based gradient computation), while Langevin dynamics (35.5\%) and denoising (11.2\%) constitute most of the remainder (PCA-based noise estimation is negligible at 1.8\%). To reduce this overhead, future work could replace backprop-based gradients with cheaper estimators such as JVP-based forward gradients\cite{baydin2018automatic,baydin2022gradients} or SPSA\cite{spall1992multivariate,spall2002implementation}.

\begin{figure}
    \centering
    \includegraphics[width=\textwidth]{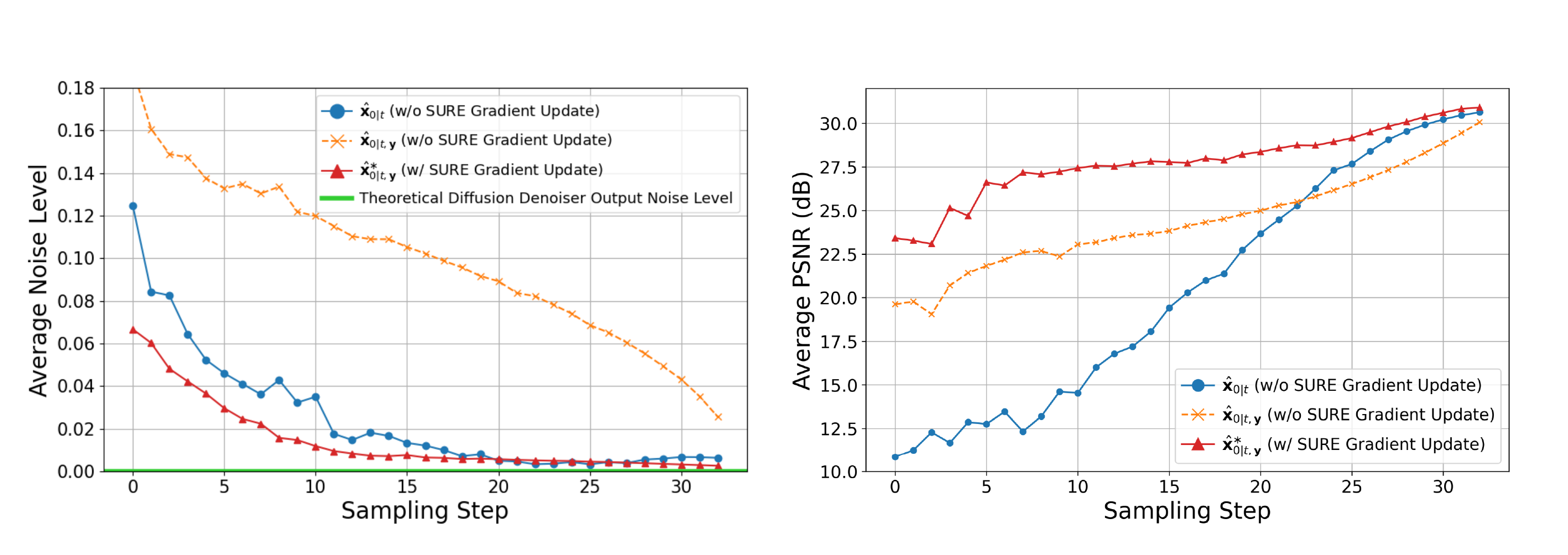}
    \caption{Average noise levels (left) and average PSNR (right) across sampling steps in SGPS with and without the SURE gradient update, evaluated on the SR\( \times 4 \) task with 33 sampling steps (T=33, NFE=99) over 100 samples, using PCA-based noise level estimation.}
    \label{noise_influx}
\end{figure}

\subsection{Ablation Study}

\paragraph{Effectiveness of SURE gradients in reducing noise levels}
To evaluate the SURE gradient update’s role in noise reduction, we measured the noise levels of \( \hat{\mathbf{x}}_{0|t} \) (denoising output) and \( \hat{\mathbf{x}}_{0|t, \mathbf{y}} \) (conditional guidance output) at each SGPS sampling step without the SURE gradient update, comparing them to \( \hat{\mathbf{x}}^{*}_{0|t, \mathbf{y}} \) with the SURE gradient update, using PCA-based noise level estimation, which has been experimentally validated as effective (Fig.~\ref{PCA} in Appendix).

As shown in Figure~\ref{noise_influx}, the plot on the left depicts the evolution of the noise level. Without SURE, the noise in \( \hat{\mathbf{x}}_{0|t, \mathbf{y}} \) exceeds that of \( \hat{\mathbf{x}}_{0|t} \) in most steps due to conditional guidance. With SURE, the noise is reduced, aligning more closely with the theoretical diffusion denoiser output.

The right plot examines the Peak Signal-to-Noise Ratio (PSNR). Without SURE, \( \hat{\mathbf{x}}_{0|t, \mathbf{y}} \) achieves a higher PSNR than \( \hat{\mathbf{x}}_{0|t} \), since the measurement \( \mathbf{y} \) provides useful reconstruction information. However, this comes with an excessive accumulation of noise in the early to mid-sampling stages. The SURE gradient update regularizes noise, reducing early-stage errors and enabling posterior sampling that better approximates the true posterior distribution, ultimately yielding final samples with higher fidelity to the ground truth.
\section{Limitation and future work}
\label{sec:Limitation and future work}

SGPS is developed for pixel-space diffusion samplers that alternate denoising and data-consistency updates. Extending it to mainstream latent diffusion model (LDM) pipelines will likely require additional design choices—such as decode--update--encode cycles~\cite{Kwon_2025_ICCV} or explicit pixel-space consistency steps—and careful validation. While our theoretical analysis suggests stable behavior under increasing dimension (under local assumptions, a small effective step size, and a Monte Carlo trace estimator for SURE that avoids explicit Jacobians), we have not exhaustively tested SGPS in extremely high-dimensional regimes beyond our current image resolutions; a systematic scaling study with higher-resolution images and larger problem sizes remains an important direction. Beyond natural image restoration, SGPS may also transfer to broader data domains where diffusion priors are effective, including medical imaging~\cite{zhu2025cross, wang2025implicit}. Finally, a key next step is extending SGPS to blind and more realistic inverse problems~\cite{liu2024adaptbir,qiao2025learning,zhao2025retinex}, where the forward operator and noise statistics are partially unknown and may deviate from idealized assumptions, motivating evaluations on real-world datasets and more flexible operator/noise modeling.

\section{Conclusion}
\label{sec:conclusion}
We proposed SGPS, a diffusion-based inverse-problem solver that mitigates sampling error accumulation by combining a SURE-driven trajectory correction with PCA-based noise-level estimation. This design provides a simple and effective mechanism to realign the sampling trajectory with the intended noise schedule, leading to high-quality reconstructions with fewer than 100 NFEs across a range of imaging tasks. In addition, our experiments show that SURE is most effective when the residual noise induced by conditional guidance is approximately Gaussian, which helps explain when and why the trajectory correction works well in practice.
\clearpage

\section*{Acknowledgments}
This work was supported in part by Korea Institute of Energy Technology Evaluation and Planning (KETEP) grant funded by the Korea Government (MOTIE) (RS-2023-00243974, Graduate School of Digital-based Sustainable Energy Process Innovation Convergence), in part by Institute of Information \& communications Technology Planning \& Evaluation (IITP) under the Leading Generative AI Human Resources Development (IITP-2025-RS-2024-00360227) grant funded by the Korea government (MSIT), and in part by the National Research Foundation of Korea (NRF) grant funded by the Korea government (MSIT) (RS-2025-24683103)

\section*{Declaration of competing interest}

The authors declare that they have no known competing financial interests or personal relationships that could have appeared to influence the work reported in this paper.

\section*{Data availability}

The FFHQ dataset used in this study is publicly available from the official project page: \url{https://github.com/NVlabs/ffhq-dataset}. We used the FFHQ images with indices \texttt{00000}--\texttt{00099}. All data were used in accordance with the dataset's license and terms of use.

\section*{Declaration of generative AI and AI-assisted technologies in the manuscript preparation process.}

During the preparation of this work the author(s) used ChatGPT in order to translate text. After using this tool/service, the author(s) reviewed and edited the content as needed and take(s) full responsibility for the content of the published article.

\section*{CRediT authorship contribution statement}

\textbf{Minwoo Kim}: Methodology, Software, Investigation, Validation, Visualization, Formal analysis, Data curation, Writing -- original draft

\textbf{Hongki Lim}: Supervision, Conceptualization, Methodology, Funding acquisition, Project administration, Resources, Writing -- review \& editing
\clearpage
\bibliographystyle{plain}
\bibliography{cas-refs}

\begin{thebibliography}{10}

\bibitem{anciukevivcius2023renderdiffusion}
Titas Anciukevi{\v{c}}ius, Zexiang Xu, Matthew Fisher, Paul Henderson, Hakan Bilen, Niloy~J Mitra, and Paul Guerrero.
\newblock Renderdiffusion: Image diffusion for 3d reconstruction, inpainting and generation.
\newblock In {\em Proceedings of the IEEE/CVF conference on computer vision and pattern recognition}, pages 12608--12618, 2023.

\bibitem{baydin2018automatic}
Atilim~Gunes Baydin, Barak~A Pearlmutter, Alexey~Andreyevich Radul, and Jeffrey~Mark Siskind.
\newblock Automatic differentiation in machine learning: a survey.
\newblock {\em Journal of machine learning research}, 18(153):1--43, 2018.

\bibitem{baydin2022gradients}
Atilim~G{\"u}nes Baydin, Barak~A Pearlmutter, Don Syme, Frank Wood, and Philip~HS Torr.
\newblock Gradients without backpropagation.
\newblock {\em CoRR}, 2022.

\bibitem{Cardoso2023MonteCG}
Gabriel Cardoso, Sylvain Le~Corff, Eric Moulines, et~al.
\newblock Monte carlo guided denoising diffusion models for bayesian linear inverse problems.
\newblock In {\em The Twelfth International Conference on Learning Representations}, 2024.

\bibitem{chen2015efficient}
Guangyong Chen, Fengyuan Zhu, and Pheng Ann~Heng.
\newblock An efficient statistical method for image noise level estimation.
\newblock In {\em Proceedings of the IEEE international conference on computer vision}, pages 477--485, 2015.

\bibitem{chungdiffusion}
Hyungjin Chung, Jeongsol Kim, Michael~Thompson Mccann, Marc~Louis Klasky, and Jong~Chul Ye.
\newblock Diffusion posterior sampling for general noisy inverse problems.
\newblock In {\em The Eleventh International Conference on Learning Representations}, 2023.

\bibitem{DDS}
Hyungjin Chung, Suhyeon Lee, and Jong-Chul Ye.
\newblock Decomposed diffusion sampler for accelerating large-scale inverse problems.
\newblock In {\em International Conference on Learning Representations}, 2023.

\bibitem{Daras2024ASO}
Giannis Daras, Hyungjin Chung, Chieh-Hsin Lai, Yuki Mitsufuji, Jong~Chul Ye, Peyman Milanfar, Alexandros~G Dimakis, and Mauricio Delbracio.
\newblock A survey on diffusion models for inverse problems.
\newblock {\em CoRR}, 2024.

\bibitem{Dou2024DiffusionPS}
Zehao Dou and Yang Song.
\newblock Diffusion posterior sampling for linear inverse problem solving: A filtering perspective.
\newblock In {\em International Conference on Learning Representations}, 2024.

\bibitem{Edupuganti2019UncertaintyQI}
Vineet Edupuganti, Morteza Mardani, Shreyas~S. Vasanawala, and John~M. Pauly.
\newblock Uncertainty quantification in deep mri reconstruction.
\newblock {\em IEEE Transactions on Medical Imaging}, 40:239--250, 2019.

\bibitem{Feng2023ScoreBasedDM}
Berthy~T. Feng, Jamie Smith, Michael Rubinstein, Huiwen Chang, Katherine~L. Bouman, and William~T. Freeman.
\newblock Score-based diffusion models as principled priors for inverse imaging.
\newblock In {\em Proceedings of the IEEE/CVF International Conference on Computer Vision (ICCV)}, pages 10520--10531, 2023.

\bibitem{he2023latentvideodiffusionmodels}
Yingqing He, Tianyu Yang, Yong Zhang, Ying Shan, and Qifeng Chen.
\newblock Latent video diffusion models for high-fidelity long video generation, 2023.

\bibitem{MPGD}
Yutong He, Naoki Murata, Chieh-Hsin Lai, Yuhta Takida, Toshimitsu Uesaka, Dongjun Kim, Wei-Hsiang Liao, Yuki Mitsufuji, J~Zico Kolter, Ruslan Salakhutdinov, et~al.
\newblock Manifold preserving guided diffusion.
\newblock In {\em The Twelfth International Conference on Learning Representations}, 2024.

\bibitem{Ho2020DenoisingDP}
Jonathan Ho, Ajay Jain, and Pieter Abbeel.
\newblock Denoising diffusion probabilistic models.
\newblock In {\em Advances in Neural Information Processing Systems}, pages 6840--6851, 2020.

\bibitem{ho2022video}
Jonathan Ho, Tim Salimans, Alexey Gritsenko, William Chan, Mohammad Norouzi, and David~J Fleet.
\newblock Video diffusion models.
\newblock In {\em Advances in Neural Information Processing Systems}, pages 8633--8646, 2022.

\bibitem{Karras2022ElucidatingTD}
Tero Karras, Miika Aittala, Timo Aila, and Samuli Laine.
\newblock Elucidating the design space of diffusion-based generative models.
\newblock In {\em Advances in Neural Information Processing Systems}, pages 26565--26577, 2022.

\bibitem{Karras2018ASG}
Tero Karras, Samuli Laine, and Timo Aila.
\newblock A style-based generator architecture for generative adversarial networks.
\newblock In {\em Proceedings of the IEEE/CVF conference on computer vision and pattern recognition}, pages 4401--4410, 2019.

\bibitem{DDRM}
Bahjat Kawar, Michael Elad, Stefano Ermon, and Jiaming Song.
\newblock Denoising diffusion restoration models.
\newblock In S.~Koyejo, S.~Mohamed, A.~Agarwal, D.~Belgrave, K.~Cho, and A.~Oh, editors, {\em Advances in Neural Information Processing Systems}, volume~35, pages 23593--23606. Curran Associates, Inc., 2022.

\bibitem{Kwon_2025_ICCV}
Taesung Kwon and Jong~Chul Ye.
\newblock Vision-xl: High definition video inverse problem solver using latent image diffusion models.
\newblock In {\em Proceedings of the IEEE/CVF International Conference on Computer Vision (ICCV)}, pages 10465--10474, 2025.

\bibitem{ling2024alignyourgaussians}
Huan Ling, Seung~Wook Kim, Antonio Torralba, Sanja Fidler, and Karsten Kreis.
\newblock Align your gaussians: Text-to-4d with dynamic 3d gaussians and composed diffusion models.
\newblock In {\em Proceedings of the IEEE/CVF conference on computer vision and pattern recognition}, pages 8576--8588, 2024.

\bibitem{Lipman2022FlowMF}
Yaron Lipman, Ricky~TQ Chen, Heli Ben-Hamu, Maximilian Nickel, and Matthew Le.
\newblock Flow matching for generative modeling.
\newblock In {\em The Eleventh International Conference on Learning Representations}, 2023.

\bibitem{liu2023zero1to3}
Ruoshi Liu, Rundi Wu, Basile Van~Hoorick, Pavel Tokmakov, Sergey Zakharov, and Carl Vondrick.
\newblock Zero-1-to-3: Zero-shot one image to 3d object.
\newblock In {\em Proceedings of the IEEE/CVF international conference on computer vision}, pages 9298--9309, 2023.

\bibitem{liu2024adaptbir}
Yingqi Liu, Jingwen He, Yihao Liu, Xinqi Lin, Fanghua Yu, Jinfan Hu, Yu~Qiao, and Chao Dong.
\newblock Adaptbir: Adaptive blind image restoration with latent diffusion prior for higher fidelity.
\newblock {\em Pattern Recognition}, 155:110659, 2024.

\bibitem{mardanivariational}
Morteza Mardani, Jiaming Song, Jan Kautz, and Arash Vahdat.
\newblock A variational perspective on solving inverse problems with diffusion models.
\newblock In {\em The Twelfth International Conference on Learning Representations}, 2024.

\bibitem{Metzler2018UnsupervisedLW}
Christopher~A. Metzler, Ali Mousavi, Reinhard Heckel, and Richard Baraniuk.
\newblock Unsupervised learning with stein's unbiased risk estimator.
\newblock {\em ArXiv}, abs/1805.10531, 2018.

\bibitem{moufad2025variational}
Badr MOUFAD, Yazid Janati, Lisa Bedin, Alain~Oliviero Durmus, randal douc, Eric Moulines, and Jimmy Olsson.
\newblock Variational diffusion posterior sampling with midpoint guidance.
\newblock In {\em The Thirteenth International Conference on Learning Representations}, 2025.

\bibitem{poole2022dreamfusion}
Ben Poole, Ajay Jain, Jonathan~T Barron, and Ben Mildenhall.
\newblock Dreamfusion: Text-to-3d using 2d diffusion.
\newblock In {\em The Eleventh International Conference on Learning Representations}, 2023.

\bibitem{qiao2025learning}
Yuanjian Qiao, Mingwen Shao, Lingzhuang Meng, and Wangmeng Zuo.
\newblock Learning physical-aware diffusion priors for zero-shot restoration of scattering-affected images.
\newblock {\em Pattern Recognition}, 163:111473, 2025.

\bibitem{Ramani2008MonteCarloSA}
Sathish Ramani, Thierry Blu, and Michael Unser.
\newblock Monte-carlo sure: A black-box optimization of regularization parameters for general denoising algorithms.
\newblock {\em IEEE Transactions on image processing}, 17(9):1540--1554, 2008.

\bibitem{rombach2021highresolution}
Robin Rombach, Andreas Blattmann, Dominik Lorenz, Patrick Esser, and Bj{\"o}rn Ommer.
\newblock High-resolution image synthesis with latent diffusion models.
\newblock In {\em Proceedings of the IEEE/CVF conference on computer vision and pattern recognition}, pages 10684--10695, 2022.

\bibitem{singermake}
Uriel Singer, Adam Polyak, Thomas Hayes, Xi~Yin, Jie An, Songyang Zhang, Qiyuan Hu, Harry Yang, Oron Ashual, Oran Gafni, et~al.
\newblock Make-a-video: Text-to-video generation without text-video data.
\newblock In {\em The Eleventh International Conference on Learning Representations}, 2022.

\bibitem{Song2020DenoisingDI}
Jiaming Song, Chenlin Meng, and Stefano Ermon.
\newblock Denoising diffusion implicit models.
\newblock In {\em International Conference on Learning Representations}, 2021.

\bibitem{song2023pseudoinverse}
Jiaming Song, Arash Vahdat, Morteza Mardani, and Jan Kautz.
\newblock Pseudoinverse-guided diffusion models for inverse problems.
\newblock In {\em International Conference on Learning Representations}, 2023.

\bibitem{Song2019GenerativeMB}
Yang Song and Stefano Ermon.
\newblock Generative modeling by estimating gradients of the data distribution.
\newblock In {\em Advances in Neural Information Processing Systems}, 2019.

\bibitem{Song2020ImprovedTF}
Yang Song and Stefano Ermon.
\newblock Improved techniques for training score-based generative models.
\newblock In {\em Advances in Neural Information Processing Systems}, pages 12438--12448, 2020.

\bibitem{Song2021SolvingIP}
Yang Song, Liyue Shen, Lei Xing, and Stefano Ermon.
\newblock Solving inverse problems in medical imaging with score-based generative models.
\newblock In {\em International Conference on Learning Representations}, 2022.

\bibitem{songscore}
Yang Song, Jascha Sohl-Dickstein, Diederik~P Kingma, Abhishek Kumar, Stefano Ermon, and Ben Poole.
\newblock Score-based generative modeling through stochastic differential equations.
\newblock In {\em International Conference on Learning Representations}, 2021.

\bibitem{spall1992multivariate}
James~C Spall.
\newblock Multivariate stochastic approximation using a simultaneous perturbation gradient approximation.
\newblock {\em IEEE transactions on automatic control}, 37(3):332--341, 1992.

\bibitem{spall2002implementation}
James~C Spall.
\newblock Implementation of the simultaneous perturbation algorithm for stochastic optimization.
\newblock {\em IEEE Transactions on aerospace and electronic systems}, 34(3):817--823, 2002.

\bibitem{charles_stein_1981}
Charles~M Stein.
\newblock Estimation of the mean of a multivariate normal distribution.
\newblock {\em The annals of Statistics}, pages 1135--1151, 1981.

\bibitem{Tran2021ExploreID}
Phong Tran, Anh~Tuan Tran, Quynh Phung, and Minh Hoai.
\newblock Explore image deblurring via encoded blur kernel space.
\newblock In {\em Proceedings of the IEEE/CVF conference on computer vision and pattern recognition}, pages 11956--11965, 2021.

\bibitem{dmplug}
Hengkang Wang, Xu~Zhang, Taihui Li, Yuxiang Wan, Tiancong Chen, and Ju~Sun.
\newblock Dmplug: A plug-in method for solving inverse problems with diffusion models.
\newblock In {\em Advances in Neural Information Processing Systems}, pages 117881--117916, 2024.

\bibitem{DDNM}
Yinhuai Wang, Jiwen Yu, and Jian Zhang.
\newblock Zero-shot image restoration using denoising diffusion null-space model.
\newblock In {\em The Eleventh International Conference on Learning Representations}, 2023.

\bibitem{wang2025implicit}
Yuang Wang, Siyeop Yoon, Pengfei Jin, Matthew Tivnan, Sifan Song, Zhennong Chen, Rui Hu, Li~Zhang, Quanzheng Li, Zhiqiang Chen, et~al.
\newblock Implicit image-to-image schr{\"o}dinger bridge for image restoration.
\newblock {\em Pattern Recognition}, 165:111627, 2025.

\bibitem{Welling2011BayesianLV}
Max Welling and Yee~W Teh.
\newblock Bayesian learning via stochastic gradient langevin dynamics.
\newblock In {\em Proceedings of the 28th international conference on machine learning (ICML-11)}, pages 681--688, 2011.

\bibitem{pnpdm}
Zihui Wu, Yu~Sun, Yifan Chen, Bingliang Zhang, Yisong Yue, and Katherine~L. Bouman.
\newblock Principled probabilistic imaging using diffusion models as plug-and-play priors.
\newblock In {\em Advances in Neural Information Processing Systems}, pages 118389--118427, 2024.

\bibitem{dpnp}
Xingyu Xu and Yuejie Chi.
\newblock Provably robust score-based diffusion posterior sampling for plug-and-play image reconstruction.
\newblock In {\em Advances in Neural Information Processing Systems}, pages 36148--36184, 2024.

\bibitem{DAPS}
Bingliang Zhang, Wenda Chu, Julius Berner, Chenlin Meng, Anima Anandkumar, and Yang Song.
\newblock Improving diffusion inverse problem solving with decoupled noise annealing.
\newblock In {\em Proceedings of the Computer Vision and Pattern Recognition Conference}, pages 20895--20905, 2025.

\bibitem{zhao2025retinex}
Zunjin Zhao and Daming Shi.
\newblock Retinex-guided generative diffusion prior for low-light image enhancement.
\newblock {\em Pattern Recognition}, page 112421, 2025.

\bibitem{zhen2025token}
Ting Zhen, Jiale Cao, Xuebin Sun, Jing Pan, Zhong Ji, and Yanwei Pang.
\newblock Token-aware and step-aware acceleration for stable diffusion.
\newblock {\em Pattern Recognition}, 164:111479, 2025.

\bibitem{zhu2025cross}
Qikui Zhu, Shaoming Zhu, Bo~Du, and Yanqing Wang.
\newblock Cross-domain distribution adversarial diffusion model for synthesizing contrast-enhanced abdomen ct imaging.
\newblock {\em Pattern Recognition}, page 111695, 2025.

\end{thebibliography}

\newpage
\appendix

\section{Theoretical Analysis}
\label{appendix:theory}
 
In this appendix, we provide detailed theoretical analysis supporting the main claims of SGPS, particularly the validity of applying SURE-based gradients to correct sampling trajectories in diffusion-based inverse problem solving.

\subsection{Gaussian Preservation Properties}
\label{appendix:gaussian_preservation}

Our theoretical analysis begins by showing that if a sample is approximately Gaussian at a given step, then applying a small-step conditional guidance update (via Langevin dynamics) preserves its near-Gaussian distribution in terms of the Wasserstein-2 distance.

\subsubsection{Preliminaries on Langevin Dynamics}

The conditional guidance step in SGPS employs Langevin dynamics, modeled by the stochastic differential equation:
\begin{equation}
  d\mathbf{x}_t \;=\; -\nabla U(\mathbf{x}_t)\,dt \;+\;\sqrt{2}\,d\mathbf{w}_t,
\end{equation}
where \(U\) is a potential function (in our case, derived from the negative log-posterior), and \(\mathbf{w}_t\) is a standard Wiener process. Its discretized form with step size \(\eta\) is:
\begin{equation}
  \mathbf{x}_{t+1} 
  \;=\; 
  \mathbf{x}_t \;-\;\eta\,\nabla U(\mathbf{x}_t)
  \;+\;\sqrt{2\,\eta}\,\boldsymbol{\xi},
  \quad
  \boldsymbol{\xi} \sim \mathcal{N}(0,\mathbf{I}),
\end{equation}
assuming that each \(\boldsymbol{\xi}\) is drawn independently at each step.

\subsubsection{Assumptions for Single-Step Gaussian Preservation}

\begin{assumption}[Locally Lipschitz Likelihood]
\label{assump:local-lipschitz-likely}
Let \(\mathcal{X}\subset\mathbb{R}^n\) be a domain containing the iterates with high probability, and for any \(\mathbf{u},\mathbf{v}\in\mathcal{X}\),
\begin{equation}
   \|\nabla\log p(\mathbf{y}\mid\mathbf{u})
    \;-\;\nabla\log p(\mathbf{y}\mid\mathbf{v})\|
   \;\leq\;
   L_\ell\,\|\mathbf{u}-\mathbf{v}\|,
\end{equation}
where \(L_\ell\) is the Lipschitz constant.
\end{assumption}

\begin{assumption}[Incoming Gaussian]
\label{assump:incoming-gaussian}
We assume \(\hat{\mathbf{x}}_{0\mid t}\sim\mathcal{N}(\mathbf{m},\sigma^2\mathbf{I})\) (or approximately) with high probability in \(\mathcal{X}\). As a consequence, \(\|\nabla\log p(\mathbf{y}\mid \hat{\mathbf{x}}_{0\mid t})\|\) is bounded by a linear expansion in terms of \(L_\ell,\sigma,\) and \(\|\nabla\log p(\mathbf{y}\mid \mathbf{m})\|\).
\end{assumption}

\subsubsection{Theorem on Single-Step Gaussian Preservation}

\begin{theorem}[Single-Step Gaussian Preservation]
\label{thm:one-step-gaussian}
Under Assumptions~\ref{assump:local-lipschitz-likely} and~\ref{assump:incoming-gaussian}, let \(\eta>0\) be small (e.g.\ \(\eta \le0.5\)). Define
\begin{equation}
   \hat{\mathbf{x}}_{0\mid t,\mathbf{y}}
   \;:=\;
   \hat{\mathbf{x}}_{0\mid t}
   \;+\;\eta\,\nabla\log p(\mathbf{y}\mid \hat{\mathbf{x}}_{0\mid t})
   \;+\;\sqrt{2\,\eta}\,\sigma\,\boldsymbol{\xi},
   \quad
   R \;:=\; \hat{\mathbf{x}}_{0\mid t,\mathbf{y}} - \mathbf{m},
\end{equation}
and let \(G_{\mathrm{target}}\sim \mathcal{N}(0,\sigma^2\mathbf{I})\).  
Then the Wasserstein-2 distance between \(\mathcal{L}(R)\) and \(G_{\mathrm{target}}\) is bounded by
\begin{equation}
  W_2^2\bigl(\mathcal{L}(R),\,G_{\mathrm{target}}\bigr)
  \;=\;
  O\bigl(\eta^2\,n\,\sigma^2\bigr).
\end{equation}
\end{theorem}

\paragraph{Proof of Theorem~\ref{thm:one-step-gaussian}.}

\paragraph{Step 1: Coupling construction.}
We define:
\begin{equation}
  R
  \;=\;
  \hat{\mathbf{x}}_{0\mid t,\mathbf{y}} - \mathbf{m}
  \;=\;
  (\hat{\mathbf{x}}_{0\mid t} - \mathbf{m})
  \;+\;\eta\,\nabla\log p(\mathbf{y}\mid\hat{\mathbf{x}}_{0\mid t})
  \;+\;\sqrt{2\,\eta}\,\sigma\,\boldsymbol{\xi}.
\end{equation}
Let 
\begin{equation}
  G_{\mathrm{coupled}}
  \;=\;
  (\hat{\mathbf{x}}_{0\mid t} - \mathbf{m})
  \;+\;\sqrt{2\,\eta}\,\sigma\,\boldsymbol{\xi}.
\end{equation}
Since \(\hat{\mathbf{x}}_{0\mid t}-\mathbf{m}\sim \mathcal{N}(0,\sigma^2\mathbf{I})\) and \(\sqrt{2\,\eta}\,\sigma\,\boldsymbol{\xi}\sim \mathcal{N}(0,2\,\eta\,\sigma^2\mathbf{I})\) are independent, \(G_{\mathrm{coupled}} \sim \mathcal{N}\bigl(0,\sigma^2(1+2\eta)\mathbf{I}\bigr)\).

\paragraph{Step 2: Measuring the coupling distance.}
\begin{equation}
  R - G_{\mathrm{coupled}}
  \;=\;
  \,\eta\,\nabla\log p(\mathbf{y}\mid \hat{\mathbf{x}}_{0\mid t}),
  \quad
  \mathbb{E}\|R - G_{\mathrm{coupled}}\|^2
  \;=\;
  \eta^2 \,\mathbb{E}\|\nabla\log p(\mathbf{y}\mid \hat{\mathbf{x}}_{0\mid t})\|^2.
\end{equation}
By Assumption~\ref{assump:local-lipschitz-likely},
\begin{equation}
  \|\nabla\log p(\mathbf{y}\mid \hat{\mathbf{x}}_{0\mid t})\|
  \;\le\;
  \|\nabla\log p(\mathbf{y}\mid \mathbf{m})\|
  + L_\ell\,\|\hat{\mathbf{x}}_{0\mid t} - \mathbf{m}\|.
\end{equation}
Since \(\hat{\mathbf{x}}_{0\mid t}-\mathbf{m}\sim\mathcal{N}(0,\sigma^2\mathbf{I})\), we have \(\mathbb{E}\|\hat{\mathbf{x}}_{0\mid t}-\mathbf{m}\|^2 = n\,\sigma^2\). Hence
\begin{equation}
  \mathbb{E}\|\nabla\log p(\mathbf{y}\mid \hat{\mathbf{x}}_{0\mid t})\|^2
  \;\le\;
  \bigl(\|\nabla\log p(\mathbf{y}\mid \mathbf{m})\|^2\bigr) + O\bigl(n\,\sigma^2\bigr).
\end{equation}
Therefore,
\begin{equation}
  \mathbb{E}\|R - G_{\mathrm{coupled}}\|^2 
  \;=\;
  O\bigl(\eta^2\,n\,\sigma^2\bigr).
\end{equation}

\paragraph{Step 3: Comparing $G_{\mathrm{coupled}}$ to $G_{\mathrm{target}}$.}
Let \(G_{\mathrm{target}} \sim \mathcal{N}(0,\sigma^2\mathbf{I})\). The Wasserstein-2 distance between isotropic Gaussians \(\mathcal{N}(0,\sigma_1^2\mathbf{I})\) and \(\mathcal{N}(0,\sigma_2^2\mathbf{I})\) is 
\(\displaystyle W_2^2 = n\,(\sigma_1 - \sigma_2)^2.\)
With \(\sigma_1 = \sigma\sqrt{1+2\,\eta}\) and \(\sigma_2=\sigma\), expanding for small \(\eta\) yields 
\begin{equation}
  W_2^2\bigl(G_{\mathrm{coupled}}, G_{\mathrm{target}}\bigr)
  \;=\;
  O\bigl(\eta^2\,n\,\sigma^2\bigr).
\end{equation}

\paragraph{Step 4: Triangle inequality for $W_2$.}
By the triangle inequality on \(W_2\):
\begin{equation}
  W_2\bigl(\mathcal{L}(R),\,G_{\mathrm{target}}\bigr)
  \;\le\;
  W_2\bigl(\mathcal{L}(R),\,G_{\mathrm{coupled}}\bigr)
  + W_2\bigl(G_{\mathrm{coupled}},\,G_{\mathrm{target}}\bigr).
\end{equation}
Squaring and applying \((a+b)^2\le2\,a^2+2\,b^2\):
\begin{equation}
  W_2^2(\mathcal{L}(R), G_{\mathrm{target}})
  \;\le\;
  2\,\mathbb{E}\|R - G_{\mathrm{coupled}}\|^2
  + 2\,W_2^2\bigl(G_{\mathrm{coupled}},G_{\mathrm{target}}\bigr)
  \;=\;
  O\bigl(\eta^2\,n\,\sigma^2\bigr).
\end{equation}
Hence we conclude
\begin{equation}
  W_2^2(\mathcal{L}(R),\,G_{\mathrm{target}})
  \;=\;
  O(\eta^2\,n\,\sigma^2).
\end{equation}
\qed

\subsubsection{Multi-Step Gaussian Preservation}

While Theorem~\ref{thm:one-step-gaussian} shows that one small-step Langevin update keeps the sample close to the original Gaussian in $W_2$, we now extend it to a sequence of $K$ such steps. Essentially, each step adds an $O(\eta^2\,n\,\sigma^2)$ error, which can accumulate over $K$ iterations.

\begin{theorem}[Multi-Step Extension]
\label{thm:multi-step-gaussian}
Suppose at step $t$ the sample $\hat{\mathbf{x}}_{0\mid t}$ is approximately $\mathcal{N}(\mathbf{m}_t,\sigma_t^2\mathbf{I})$, and we apply a small-step Langevin update
\begin{equation}
  \hat{\mathbf{x}}_{0\mid t,\mathbf{y}}
  \;=\;
  \hat{\mathbf{x}}_{0\mid t}
  \;+\;
  \eta\,\nabla\log p(\mathbf{y}\mid\hat{\mathbf{x}}_{0\mid t})
  \;+\;\sqrt{2\,\eta}\,\sigma_t\,\boldsymbol{\xi}.
\end{equation}
If this process is repeated over $K$ total steps (with potentially varying $\sigma_t$ and small $\eta$ each time), then for each intermediate state,
\begin{equation}
  W_2^2\!\Bigl(\mathcal{L}\!\bigl(\hat{\mathbf{x}}_{0\mid t-k,\mathbf{y}} - \mathbf{m}_{t-k}\bigr),
  \,\mathcal{N}(0,\sigma_{t-k}^2\,\mathbf{I})\Bigr)
  \;=\;
  O\!\bigl(K\,\eta^2\,n\,\max_{1\le i \le K}\{\sigma_{t-i}^2\}\bigr).
\end{equation}
In other words, the $W_2$ deviation accumulates at most linearly in $K$, as long as each per-step update is small.
\end{theorem}

\paragraph{Proof.}
One can argue inductively:
\begin{itemize}
\item[(i)] Apply Theorem~\ref{thm:one-step-gaussian} at each step $k=1,\ldots,K$.  
\item[(ii)] Each step contributes an $O(\eta^2\,n\,\sigma_{t-k}^2)$ deviation in $W_2^2$.  
\item[(iii)] Summing or bounding across $K$ updates yields $K$ times that base order, giving $O(K\,\eta^2\,n\,\max\sigma_{t-k}^2)$.  
\end{itemize}

When combined with SGPS corrections after each measurement-consistency update, the sample mean $\mathbf{m}_{t-k}$ remains close to the true manifold, and the variance remains close to the prescribed $\sigma_{t-k}^2$. This justifies that repeated small-step Langevin guidance does not drastically increase deviations from an approximate Gaussian structure, even over multiple steps.
\qed

\subsection{SURE-Based Gradients and KL Convergence}
\label{appendix:sure_gradients}
 
In this section, we establish how SURE-based gradients, under suitable conditions, approximate those of the posterior distribution and enable convergence in the KL divergence sense.
 
\subsubsection{Background on SURE}
 
For a denoiser function $D$ applied to noisy data $\mathbf{x}_{\text{noisy}} = \mathbf{x}_0 + \mathbf{z}$ where $\mathbf{z} \sim \mathcal{N}(0, \sigma^2\mathbf{I})$, Stein's Unbiased Risk Estimate (SURE) provides an unbiased estimate of the expected mean squared error (MSE) without requiring access to the ground truth:
\begin{equation}
    \mathrm{SURE} = -n\sigma^2 + \|\mathbf{x}_{\text{noisy}} - D(\mathbf{x}_{\text{noisy}}, \sigma)\|^2 + 2\sigma^2\mathrm{div}_{\mathbf{x}_{\text{noisy}}}\{D(\mathbf{x}_{\text{noisy}}, \sigma)\}
\end{equation}
where $n$ is the dimension of $\mathbf{x}$ and $\mathrm{div}_{\mathbf{x}_{\text{noisy}}}\{D\}$ is the divergence of $D$ with respect to its input.
 
The Monte Carlo approximation of the divergence term is:
\begin{equation}
    \mathrm{div}_{\mathbf{x}_{\text{noisy}}}\{D\} \approx \frac{\mathbf{b}^T[D(\mathbf{x}_{\text{noisy}} + \epsilon\mathbf{b}, \sigma) - D(\mathbf{x}_{\text{noisy}}, \sigma)]}{\epsilon}
\end{equation}
where $\mathbf{b} \sim \mathcal{N}(0, \mathbf{I})$ and $\epsilon$ is a small constant.
 
\subsubsection{Assumptions for KL Convergence Analysis}
 
\begin{assumption}[Local Strong Convexity \& Smoothness of Negative Log Posterior]
\label{assump:strong-convex}
We define $\Phi(\mathbf{x})=-\log p(\mathbf{x}|\mathbf{y})$ and assume that it satisfies:
\begin{enumerate}
  \item $\mu$-strong convexity: $\nabla^2\Phi(\mathbf{x})\succeq \mu \mathbf{I}$ for some $\mu > 0$
  \item $L$-smoothness: $\|\nabla\Phi(\mathbf{x})-\nabla\Phi(\mathbf{z})\|\leq L\|\mathbf{x}-\mathbf{z}\|$ for some $L > 0$
\end{enumerate}
in a domain $\mathcal{X}\subset\mathbb{R}^d$ containing the iterates with high probability.
\end{assumption}
 
\begin{assumption}[Approx. Unbiased SURE Correction for Posterior Gradient]
\label{assump:sure-bias}
We define the SURE-based correction vector in SGPS as:
\begin{equation}
  g_{\mathrm{corr}}(\hat{\mathbf{x}}_{0|t,\mathbf{y}})
  :=
  \alpha\,\nabla_{\hat{\mathbf{x}}_{0|t,\mathbf{y}}}\mathrm{SURE}(t),
\end{equation}
where $\alpha$ is the algorithm's SURE step size. We assume this gradient approximates the scaled negative log-posterior gradient with bounded bias and variance:
\begin{equation}
  \mathbb{E}[g_{\mathrm{corr}}(\hat{\mathbf{x}}_{0|t,\mathbf{y}})|\hat{\mathbf{x}}_{0|t,\mathbf{y}}]
  = \hat{\sigma}_0^2 \,\nabla\Phi(\hat{\mathbf{x}}_{0|t,\mathbf{y}})
    + \varepsilon(\hat{\mathbf{x}}_{0|t,\mathbf{y}}),
\end{equation}
\begin{equation}
  \mathbb{E}\|g_{\mathrm{corr}}(\hat{\mathbf{x}}_{0|t,\mathbf{y}}) - \hat{\sigma}_0^2 \nabla\Phi(\hat{\mathbf{x}}_{0|t,\mathbf{y}})\|^2
  \leq \hat{\sigma}_0^4\,V + \varepsilon_{\mathrm{var}}(\hat{\mathbf{x}}_{0|t,\mathbf{y}}),
\end{equation}
where $\varepsilon(\hat{\mathbf{x}}_{0|t,\mathbf{y}})$ is a bias term, $\varepsilon_{\mathrm{var}}(\hat{\mathbf{x}}_{0|t,\mathbf{y}})$ is a variance expansion term, and $V$ is a constant.
\end{assumption}
 
\begin{assumption}[Scaled Step Size in the Update]
\label{assump:scaled-step-size}
Let $\beta_t = \alpha \hat{\sigma}_0^2$ be the effective step size for the gradient update, where $\alpha$ is the raw step size for $g_{\mathrm{corr}}(\hat{\mathbf{x}}_{0|t,\mathbf{y}})$. We assume $\beta_t$ is sufficiently small, specifically $\beta_t \leq \frac{1}{2L}$, where $L$ is the smoothness constant from Assumption \ref{assump:strong-convex}.
\end{assumption}
 
\subsubsection{Theorem on KL Convergence under SURE-Based Updates}
 
\begin{theorem}[KL Convergence under Biased SURE Gradients]
\label{thm:multistep-sgps-2}
Under Assumptions \ref{assump:strong-convex}, \ref{assump:sure-bias}, and \ref{assump:scaled-step-size}, let
\begin{equation}
  g(\hat{\mathbf{x}}_{0|t,\mathbf{y}}) := \frac{1}{\hat{\sigma}_0^2} g_{\mathrm{corr}}(\hat{\mathbf{x}}_{0|t,\mathbf{y}}),
  \quad
  \hat{\mathbf{x}}^{*}_{0|t,\mathbf{y}} := \hat{\mathbf{x}}_{0|t,\mathbf{y}} - \beta_t\,g(\hat{\mathbf{x}}_{0|t,\mathbf{y}}),
  \quad
  \beta_t=\alpha \hat{\sigma}_0^2.
\end{equation}
Let $q_t,q_t^*$ be the distributions of $\hat{\mathbf{x}}_{0|t,\mathbf{y}}$ before/after this update. Then:
 
\paragraph{(a) Single-step bound.}
There exist constants $C>0$ and a mismatch term $\Delta_t$ such that:
\begin{equation}
  D_{\mathrm{KL}}(q_t^*\,\|p(\mathbf{x}|\mathbf{y}))
  \leq
  (1-\beta_t\mu)\,D_{\mathrm{KL}}(q_t\|p(\mathbf{x}|\mathbf{y}))
  + \beta_t^2\,C
  + \Delta_t.
\end{equation}
 
\paragraph{(b) Multi-step telescoping.}
\begin{equation}
  D_{\mathrm{KL}}(q_1^*\|p(\mathbf{x}|\mathbf{y}))
  \leq
  \left(\prod_{k=1}^T (1-\beta_k\mu)\right)D_{\mathrm{KL}}(q_T\|p(\mathbf{x}|\mathbf{y}))
  + \sum_{k=1}^T
     \left(\prod_{\ell=k+1}^T(1-\beta_\ell\mu)\right)
     [\beta_k^2\,C + \Delta_k].
\end{equation}
\end{theorem}
 
\paragraph{Proof of Theorem \ref{thm:multistep-sgps-2}.}
 
\paragraph{Step 1: Defining the scaled gradient.}
Let $g(\hat{\mathbf{x}}_{0|t,\mathbf{y}}) = \frac{1}{\hat{\sigma}_0^2}g_{\mathrm{corr}}(\hat{\mathbf{x}}_{0|t,\mathbf{y}})$. Then:
\begin{align}
    \mathbb{E}[g(\hat{\mathbf{x}}_{0|t,\mathbf{y}})|\hat{\mathbf{x}}_{0|t,\mathbf{y}}] &= \nabla\Phi(\hat{\mathbf{x}}_{0|t,\mathbf{y}}) + \frac{1}{\hat{\sigma}_0^2}\varepsilon(\hat{\mathbf{x}}_{0|t,\mathbf{y}})\\
    \mathbb{E}\|g(\hat{\mathbf{x}}_{0|t,\mathbf{y}}) - \nabla\Phi(\hat{\mathbf{x}}_{0|t,\mathbf{y}})\|^2 &\leq V + \frac{\varepsilon_{\mathrm{var}}(\hat{\mathbf{x}}_{0|t,\mathbf{y}})}{\hat{\sigma}_0^4}
\end{align}
 
\paragraph{Step 2: Single-step $L$-smoothness expansion.}
Consider the update $\hat{\mathbf{x}}^*_{0|t,\mathbf{y}} = \hat{\mathbf{x}}_{0|t,\mathbf{y}} - \beta_t g(\hat{\mathbf{x}}_{0|t,\mathbf{y}})$ with $\beta_t = \alpha\hat{\sigma}_0^2$.
 
By the $L$-smoothness of $\Phi$ (Assumption \ref{assump:strong-convex}):
\begin{align}
    \Phi(\hat{\mathbf{x}}^*_{0|t,\mathbf{y}}) &\leq \Phi(\hat{\mathbf{x}}_{0|t,\mathbf{y}}) - \beta_t\langle\nabla\Phi(\hat{\mathbf{x}}_{0|t,\mathbf{y}}), g(\hat{\mathbf{x}}_{0|t,\mathbf{y}})\rangle + \frac{L\beta_t^2}{2}\|g(\hat{\mathbf{x}}_{0|t,\mathbf{y}})\|^2
\end{align}
 
Taking expectation:
\begin{align}
    \mathbb{E}[\Phi(\hat{\mathbf{x}}^*_{0|t,\mathbf{y}})] &\leq \mathbb{E}[\Phi(\hat{\mathbf{x}}_{0|t,\mathbf{y}})] - \beta_t\mathbb{E}[\langle\nabla\Phi(\hat{\mathbf{x}}_{0|t,\mathbf{y}}), g(\hat{\mathbf{x}}_{0|t,\mathbf{y}})\rangle] + \frac{L\beta_t^2}{2}\mathbb{E}[\|g(\hat{\mathbf{x}}_{0|t,\mathbf{y}})\|^2]
\end{align}
 
\paragraph{Step 3: Decomposing the inner product term.}
\begin{align}
    \mathbb{E}[\langle\nabla\Phi(\hat{\mathbf{x}}_{0|t,\mathbf{y}}), g(\hat{\mathbf{x}}_{0|t,\mathbf{y}})\rangle] &= \mathbb{E}[\langle\nabla\Phi(\hat{\mathbf{x}}_{0|t,\mathbf{y}}), \mathbb{E}[g(\hat{\mathbf{x}}_{0|t,\mathbf{y}})|\hat{\mathbf{x}}_{0|t,\mathbf{y}}]\rangle]\\
    &= \mathbb{E}[\langle\nabla\Phi(\hat{\mathbf{x}}_{0|t,\mathbf{y}}), \nabla\Phi(\hat{\mathbf{x}}_{0|t,\mathbf{y}}) + \frac{1}{\hat{\sigma}_0^2}\varepsilon(\hat{\mathbf{x}}_{0|t,\mathbf{y}})\rangle]\\
    &= \mathbb{E}[\|\nabla\Phi(\hat{\mathbf{x}}_{0|t,\mathbf{y}})\|^2] + \frac{1}{\hat{\sigma}_0^2}\mathbb{E}[\langle\nabla\Phi(\hat{\mathbf{x}}_{0|t,\mathbf{y}}), \varepsilon(\hat{\mathbf{x}}_{0|t,\mathbf{y}})\rangle]
\end{align}
 
\paragraph{Step 4: Bounding the gradient norm squared.}
Using the expanded form of $g(\hat{\mathbf{x}}_{0|t,\mathbf{y}})$:
\begin{align}
    g(\hat{\mathbf{x}}_{0|t,\mathbf{y}}) &= \frac{1}{\hat{\sigma}_0^2}g_{\mathrm{corr}}(\hat{\mathbf{x}}_{0|t,\mathbf{y}})\\
    &= \frac{1}{\hat{\sigma}_0^2}(g_{\mathrm{corr}}(\hat{\mathbf{x}}_{0|t,\mathbf{y}}) - \hat{\sigma}_0^2\nabla\Phi(\hat{\mathbf{x}}_{0|t,\mathbf{y}})) + \nabla\Phi(\hat{\mathbf{x}}_{0|t,\mathbf{y}})
\end{align}
 
Using the inequality $\|\mathbf{a} + \mathbf{b}\|^2 \leq 2\|\mathbf{a}\|^2 + 2\|\mathbf{b}\|^2$:
\begin{align}
    \|g(\hat{\mathbf{x}}_{0|t,\mathbf{y}})\|^2 &\leq 2\Big\|\frac{1}{\hat{\sigma}_0^2}(g_{\mathrm{corr}}(\hat{\mathbf{x}}_{0|t,\mathbf{y}}) - \hat{\sigma}_0^2\nabla\Phi(\hat{\mathbf{x}}_{0|t,\mathbf{y}}))\Big\|^2 + 2\|\nabla\Phi(\hat{\mathbf{x}}_{0|t,\mathbf{y}})\|^2
\end{align}
 
By our assumption on the variance:
\begin{align}
    \mathbb{E}\|g(\hat{\mathbf{x}}_{0|t,\mathbf{y}})\|^2 &\leq \frac{2}{\hat{\sigma}_0^4}(\hat{\sigma}_0^4V + \varepsilon_{\mathrm{var}}(\hat{\mathbf{x}}_{0|t,\mathbf{y}})) + 2\mathbb{E}\|\nabla\Phi(\hat{\mathbf{x}}_{0|t,\mathbf{y}})\|^2\\
    &= 2V + \frac{2\varepsilon_{\mathrm{var}}(\hat{\mathbf{x}}_{0|t,\mathbf{y}})}{\hat{\sigma}_0^4} + 2\mathbb{E}\|\nabla\Phi(\hat{\mathbf{x}}_{0|t,\mathbf{y}})\|^2
\end{align}
 
\paragraph{Step 5: Using $\mu$-strong convexity.}
By Assumption \ref{assump:strong-convex}, $\Phi$ is $\mu$-strongly convex, which implies:
\begin{equation}
    \|\nabla\Phi(\mathbf{x})\|^2 \geq 2\mu(\Phi(\mathbf{x}) - \Phi^*)
\end{equation}
where $\Phi^* = \min_{\mathbf{x}} \Phi(\mathbf{x})$ is the minimum value of $\Phi$.
 
Let's examine the combination of gradient terms:
\begin{align}
    -\beta_t\mathbb{E}[\|\nabla\Phi(\hat{\mathbf{x}}_{0|t,\mathbf{y}})\|^2] + L\beta_t^2\mathbb{E}[\|\nabla\Phi(\hat{\mathbf{x}}_{0|t,\mathbf{y}})\|^2] &= \mathbb{E}[\|\nabla\Phi(\hat{\mathbf{x}}_{0|t,\mathbf{y}})\|^2](-\beta_t + L\beta_t^2)
\end{align}
 
If $\beta_t \leq \frac{1}{2L}$ (as per Assumption \ref{assump:scaled-step-size}), then $-\beta_t + L\beta_t^2 \leq -\frac{\beta_t}{2}$, giving:
\begin{align}
    -\frac{\beta_t}{2}\mathbb{E}[\|\nabla\Phi(\hat{\mathbf{x}}_{0|t,\mathbf{y}})\|^2] &\leq -\beta_t\mu\mathbb{E}[\Phi(\hat{\mathbf{x}}_{0|t,\mathbf{y}}) - \Phi^*]
\end{align}
 
\paragraph{Step 6: Combining all terms.}
Gathering all terms and using $\Delta_t$ to represent the accumulated error terms:
\begin{align}
    \mathbb{E}[\Phi(\hat{\mathbf{x}}^*_{0|t,\mathbf{y}})] - \Phi^* &\leq (1-\beta_t\mu)(\mathbb{E}[\Phi(\hat{\mathbf{x}}_{0|t,\mathbf{y}})] - \Phi^*)\\
    &\quad - \frac{\beta_t}{\hat{\sigma}_0^2}\mathbb{E}[\langle\nabla\Phi(\hat{\mathbf{x}}_{0|t,\mathbf{y}}), \varepsilon(\hat{\mathbf{x}}_{0|t,\mathbf{y}})\rangle]\\
    &\quad + \frac{L\beta_t^2}{2}(2V + \frac{2\varepsilon_{\mathrm{var}}(\hat{\mathbf{x}}_{0|t,\mathbf{y}})}{\hat{\sigma}_0^4})\\
    &= (1-\beta_t\mu)(\mathbb{E}[\Phi(\hat{\mathbf{x}}_{0|t,\mathbf{y}})] - \Phi^*) + \beta_t^2LV + \Delta_t
\end{align}
where $\Delta_t$ encapsulates all the bias and variance error terms.
 
\paragraph{Step 7: Connecting to KL divergence.}
For probability distributions that are close to their stationary distribution, the KL divergence can be approximately related to the expectation of the negative log-density:
\begin{equation}
    D_{\mathrm{KL}}(q\|p(\mathbf{x}|\mathbf{y})) \approx \mathbb{E}_q[\Phi(\mathbf{x})] - \Phi^*
\end{equation}
 
Under this approximation, we can restate our result in terms of KL divergence:
\begin{equation}
    D_{\mathrm{KL}}(q_t^*\|p(\mathbf{x}|\mathbf{y})) \leq (1-\beta_t\mu)D_{\mathrm{KL}}(q_t\|p(\mathbf{x}|\mathbf{y})) + \beta_t^2C + \Delta_t
\end{equation}
where $C$ is a constant (here, $C = LV$), and $q_t, q_t^*$ are the distributions of $\hat{\mathbf{x}}_{0|t,\mathbf{y}}$ and $\hat{\mathbf{x}}^*_{0|t,\mathbf{y}}$, respectively.
 
\paragraph{Step 8: Multi-step telescoping.}
For a sequence of updates from $t=T$ to $t=1$, we can telescope the inequalities:
\begin{align}
    D_{\mathrm{KL}}(q_1^*\|p(\mathbf{x}|\mathbf{y})) &\leq (1-\beta_1\mu)D_{\mathrm{KL}}(q_1\|p(\mathbf{x}|\mathbf{y})) + \beta_1^2C + \Delta_1\\
    &\leq (1-\beta_1\mu)[(1-\beta_2\mu)D_{\mathrm{KL}}(q_2\|p(\mathbf{x}|\mathbf{y})) + \beta_2^2C + \Delta_2] + \beta_1^2C + \Delta_1\\
    &= (1-\beta_1\mu)(1-\beta_2\mu)D_{\mathrm{KL}}(q_2\|p(\mathbf{x}|\mathbf{y}))\\
    &\quad + (1-\beta_1\mu)(\beta_2^2C + \Delta_2) + \beta_1^2C + \Delta_1
\end{align}
 
Continuing this telescoping process yields:
\begin{align}
    D_{\mathrm{KL}}(q_1^*\|p(\mathbf{x}|\mathbf{y})) &\leq \left(\prod_{k=1}^T(1-\beta_k\mu)\right)D_{\mathrm{KL}}(q_T\|p(\mathbf{x}|\mathbf{y}))\\
    &\quad + \sum_{k=1}^T\left(\prod_{\ell=1}^{k-1}(1-\beta_\ell\mu)\right)(\beta_k^2C + \Delta_k)
\end{align}
 
After reindexing to match the notation in the theorem statement:
\begin{align}
    D_{\mathrm{KL}}(q_1^*\|p(\mathbf{x}|\mathbf{y})) &\leq \left(\prod_{k=1}^T(1-\beta_k\mu)\right)D_{\mathrm{KL}}(q_T\|p(\mathbf{x}|\mathbf{y}))\\
    &\quad + \sum_{k=1}^T\left(\prod_{\ell=k+1}^T(1-\beta_\ell\mu)\right)(\beta_k^2C + \Delta_k)
\end{align}
 
This completes the proof of Theorem \ref{thm:multistep-sgps-2}.
\qed

\section{SURE Derivation and Proof}
\label{appendix:sure_derivation}

In this section, we derive Stein's Unbiased Risk Estimate (SURE)~\cite{charles_stein_1981} and provide a proof for the Monte Carlo SURE (MC-SURE) approximation used in the SGPS algorithm.

\subsection{SURE Derivation}
Given a ground truth image \(\mathbf{x}_0\), the noisy image can be formulated as
\begin{equation}
\mathbf{x}_{\text{noisy}} = \mathbf{x}_0 + \mathbf{z},
\end{equation}
where \(\mathbf{z} \sim \mathcal{N}(0, \sigma^2 \mathbf{I})\) represents Gaussian noise. Let the reconstruction be \(\hat{\mathbf{x}} = f(\mathbf{x}_{\text{noisy}})\), where \(f\) is the denoiser function. The mean squared error (MSE) is defined as
\begin{equation}
\text{MSE} = \mathbb{E}\left[ \|\mathbf{x}_0 - \hat{\mathbf{x}}\|^2 \right].
\end{equation}
Expanding the MSE, we obtain
\begin{equation}
\begin{aligned}
\text{MSE} &= \mathbb{E}\left[ \|\mathbf{x}_0 - \mathbf{x}_{\text{noisy}} + \mathbf{x}_{\text{noisy}} - \hat{\mathbf{x}}\|^2 \right] \\
&= \mathbb{E}\left[ \|\mathbf{x}_0 - \mathbf{x}_{\text{noisy}}\|^2 \right] + \mathbb{E}\left[ \|\mathbf{x}_{\text{noisy}} - \hat{\mathbf{x}}\|^2 \right] + 2\mathbb{E}\left[ (\mathbf{x}_0 - \mathbf{x}_{\text{noisy}})^T (\mathbf{x}_{\text{noisy}} - \hat{\mathbf{x}}) \right] \\
&= n\sigma^2 + \mathbb{E}\left[ \|\mathbf{x}_{\text{noisy}} - \hat{\mathbf{x}}\|^2 \right] + 2 \left( -n\sigma^2 + \sum_{i=1}^n \text{Cov}(\mathbf{x}_{\text{noisy}, i}, \hat{\mathbf{x}}_i) \right) \\
&= -n\sigma^2 + \mathbb{E}\left[ \|\mathbf{x}_{\text{noisy}} - \hat{\mathbf{x}}\|^2 \right] + 2 \sum_{i=1}^n \text{Cov}(\mathbf{x}_{\text{noisy}, i}, \hat{\mathbf{x}}_i).
\end{aligned}
\end{equation}
To handle the covariance term, we apply Stein's lemma. Let \(\mathbf{X} \sim \mathcal{N}(\boldsymbol{\mu}, \sigma^2 \mathbf{I})\) be an \(n\)-dimensional normal variate, and let \(f: \mathbb{R}^n \to \mathbb{R}^n\) be an almost differentiable function satisfying \(\mathbb{E}\left[ \|f(\mathbf{X})\|_2 \right] < \infty\). Stein's lemma states
\begin{equation}
\frac{1}{\sigma^2} \mathbb{E}\left[ (\mathbf{X} - \boldsymbol{\mu}) f(\mathbf{X}) \right] = \mathbb{E}\left[ \nabla f(\mathbf{X}) \right].
\end{equation}
Applying this component-wise and summing over all indices \(i = 1, \ldots, n\), we get
\begin{equation}
\frac{1}{\sigma^2} \sum_{i=1}^n \text{Cov}(\mathbf{X}_i, f_i(\mathbf{X})) = \frac{1}{\sigma^2} \sum_{i=1}^n \mathbb{E}\left[ (\mathbf{X}_i - \mu_i) f_i(\mathbf{X}) \right] = \mathbb{E}\left[ \sum_{i=1}^n \frac{\partial f_i}{\partial \mathbf{X}_i}(\mathbf{X}) \right].
\end{equation}
In our context, \(\mathbf{X} = \mathbf{x}_{\text{noisy}}\), \(\boldsymbol{\mu} = \mathbf{x}_0\), and \(f(\mathbf{x}_{\text{noisy}}) = \hat{\mathbf{x}}\). Thus,
\begin{equation}
\sum_{i=1}^n \text{Cov}(\mathbf{x}_{\text{noisy}, i}, \hat{\mathbf{x}}_i) = \sigma^2 \mathbb{E}\left[ \sum_{i=1}^n \frac{\partial \hat{\mathbf{x}}_i}{\partial \mathbf{x}_{\text{noisy}, i}} \right] = \sigma^2 \mathbb{E}\left[ \text{div}_{\mathbf{x}_{\text{noisy}}} \{\hat{\mathbf{x}}\} \right].
\end{equation}
Substituting into the MSE expression, we obtain
\begin{equation}
\text{MSE} = -n\sigma^2 + \mathbb{E}\left[ \|\mathbf{x}_{\text{noisy}} - \hat{\mathbf{x}}\|^2 \right] + 2\sigma^2 \mathbb{E}\left[ \text{div}_{\mathbf{x}_{\text{noisy}}} \{\hat{\mathbf{x}}\} \right].
\end{equation}
Rearranging, the SURE estimator is
\begin{equation}
\text{SURE} = -n\sigma^2 + \|\mathbf{x}_{\text{noisy}} - \hat{\mathbf{x}}\|_2^2 + 2\sigma^2 \text{div}_{\mathbf{x}_{\text{noisy}}} \{\hat{\mathbf{x}}\},
\end{equation}
which is unbiased, i.e., \(\mathbb{E}[\text{SURE}] = \text{MSE}\).

\subsection{MC-SURE Theorem \texorpdfstring{\cite{Ramani2008MonteCarloSA}}{Ramani2008}}
\label{sec:mc_sure_theorem}
Computing the divergence \(\text{div}_{\mathbf{x}_{\text{noisy}}} \{\hat{\mathbf{x}}\}\) directly is challenging. The Monte Carlo SURE (MC-SURE) approach estimates it using
\begin{equation}
\text{div}_{\mathbf{y}}\{f(\mathbf{y})\} = \lim_{\epsilon \to 0} \mathbb{E}_{\mathbf{b}'} \left[ \frac{\mathbf{b}'^T \left( f(\mathbf{y} + \epsilon \mathbf{b}') - f(\mathbf{y}) \right)}{\epsilon} \right],
\end{equation}
where \(\mathbf{b}' \sim \mathcal{N}(0, \mathbf{I})\) is a zero-mean i.i.d. random vector with unit variance and bounded higher-order moments, independent of \(\mathbf{y}\). This holds assuming \(f\) has a well-defined second-order Taylor expansion or satisfies a tempered growth condition, i.e., \(\|f(\mathbf{y})\| \leq C_0 (1 + \|\mathbf{y}\|^{\eta_0})\), where \(\eta_0 > 1\) and \(C_0 > 0\).

\paragraph{Proof of Theorem \ref{sec:mc_sure_theorem}.}
\label{sec:mc_sure_proof}
Consider the Taylor expansion of \(f(\mathbf{y} + \epsilon \mathbf{b}')\) around \(\mathbf{y}\):
\begin{equation}
f(\mathbf{y} + \epsilon \mathbf{b}') = f(\mathbf{y}) + \epsilon \mathbf{J}_f(\mathbf{y}) \mathbf{b}' + \epsilon^2 \mathbf{r}_f,
\end{equation}
where \(\mathbf{J}_f(\mathbf{y})\) is the Jacobian of \(f\) at \(\mathbf{y}\), and \(\mathbf{r}_f\) is the vector of remainder terms in Lagrange form, with \(\mathbb{E}_{\mathbf{b}'}\left[ \mathbf{r}_{f,k}^2 \right] < \infty\) for each component \(k = 1, \ldots, N\). Compute the difference:
\begin{equation}
f(\mathbf{y} + \epsilon \mathbf{b}') - f(\mathbf{y}) = \epsilon \mathbf{J}_f(\mathbf{y}) \mathbf{b}' + \epsilon^2 \mathbf{r}_f.
\end{equation}
Premultiply by \(\mathbf{b}'^T\) and take the expectation over \(\mathbf{b}'\):
\begin{equation}
\mathbb{E}_{\mathbf{b}'}\left[ \mathbf{b}'^T \left( f(\mathbf{y} + \epsilon \mathbf{b}') - f(\mathbf{y}) \right) \right] = \mathbb{E}_{\mathbf{b}'}\left[ \mathbf{b}'^T \left( \epsilon \mathbf{J}_f(\mathbf{y}) \mathbf{b}' + \epsilon^2 \mathbf{r}_f \right) \right].
\end{equation}
This splits into
\begin{equation}
\epsilon \mathbb{E}_{\mathbf{b}'}\left[ \mathbf{b}'^T \mathbf{J}_f(\mathbf{y}) \mathbf{b}' \right] + \epsilon^2 \mathbb{E}_{\mathbf{b}'}\left[ \mathbf{b}'^T \mathbf{r}_f \right].
\end{equation}
Since \(\mathbf{b}'\) is i.i.d. with zero mean and unit variance, \(\mathbb{E}_{\mathbf{b}'}\left[ \mathbf{b}'^T \mathbf{J}_f(\mathbf{y}) \mathbf{b}' \right] = \text{trace}\{\mathbf{J}_f(\mathbf{y})\}\). For the remainder, let \(C_2 = \mathbb{E}_{\mathbf{b}'}\left[ \mathbf{b}'^T \mathbf{r}_f \right]\), where \(|C_2| < \infty\) due to the boundedness of \(\mathbf{r}_{f,k}\). Thus,
\begin{equation}
\mathbb{E}_{\mathbf{b}'}\left[ \mathbf{b}'^T \left( f(\mathbf{y} + \epsilon \mathbf{b}') - f(\mathbf{y}) \right) \right] = \epsilon \text{trace}\{\mathbf{J}_f(\mathbf{y})\} + \epsilon^2 C_2.
\end{equation}
Dividing by \(\epsilon\) and taking the limit as \(\epsilon \to 0\), the second term vanishes:
\begin{equation}
\lim_{\epsilon \to 0} \frac{1}{\epsilon} \mathbb{E}_{\mathbf{b}'}\left[ \mathbf{b}'^T \left( f(\mathbf{y} + \epsilon \mathbf{b}') - f(\mathbf{y}) \right) \right] = \text{trace}\{\mathbf{J}_f(\mathbf{y})\} = \text{div}_{\mathbf{y}}\{f(\mathbf{y})\}.
\end{equation}
In practice, due to finite machine precision, MC-SURE approximates
\begin{equation}
\text{div}_{\mathbf{y}}\{f(\mathbf{y})\} \approx \frac{\mathbf{b}'^T \left( f(\mathbf{y} + \epsilon \mathbf{b}') - f(\mathbf{y}) \right)}{\epsilon},
\end{equation}
where \(\epsilon\) is chosen to balance approximation accuracy and numerical stability.
\qed

\clearpage
\begin{figure}
    \centering
    \includegraphics[width=0.8\linewidth]{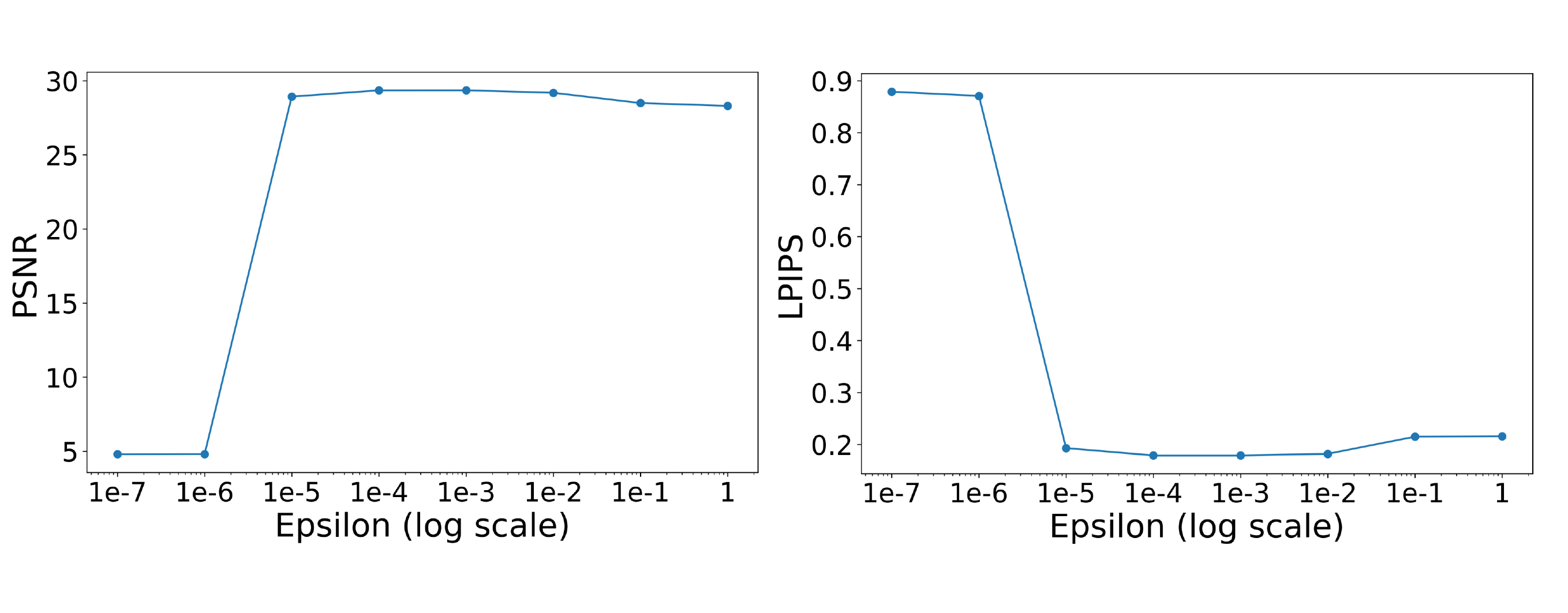}
    \caption{Sensitivity Analysis of the MC-SURE Perturbation Scale ($\epsilon$) on PSNR and LPIPS Performance for the SR4 Task.}
    \label{epsilon}
\end{figure}

\section{Noise Schedule and Hyper-Parameter Analysis}
We utilize the VP-preconditioned DDPM denoiser of \cite{Karras2022ElucidatingTD} and the noise scheduler \( \sigma_t = t \) in the diffusion denoising step. The noise scheduling approach in \cite{Karras2022ElucidatingTD} ensures that as \( \sigma_t \) decreases, the step size monotonically decreases, facilitating more frequent sampling in regions with lower noise. This strategy is adopted because errors in low-noise regions have a larger impact and, by increasing sampling frequency in these regions, it reduces errors and ultimately ensures that the samples follow the solution trajectory more closely. We use \( \rho = 7 \), \( t_{\min} = 0.02 \), and \( t_{\max} = T \). 
\begin{equation}
t_i = \left( t_{\text{max}}^{\frac{1}{\rho}} + \frac{i}{N-1} \left( t_{\text{min}}^{\frac{1}{\rho}} - t_{\text{max}}^{\frac{1}{\rho}} \right) \right)^\rho
\end{equation}
In the SURE gradient update, we use the following parameters: \(\epsilon = \frac{\max(\mathbf{x}_{\text{noisy}})}{1000}\), \(\mathbf{b}\) is sampled from \(N(0,1)\), \(n = 3 \times 256 \times 256\), and \(\alpha = 0.5\). A detailed analysis and selection of the hyperparameters is provided in the following.

\subsection{Hyperparameter \texorpdfstring{$\epsilon$}{epsilon}}
As discussed in the previous section, one of the conditions of the MC-SURE Theorem is that \(\epsilon\) must be sufficiently small. However, if \(\epsilon\) becomes excessively small, round-off errors may arise, necessitating the selection of an appropriately balanced value. According to MC-SURE, the lower bound of \(\epsilon\) depends on the sensitivity of the function \(f\). In our study, where \(f\) represents a diffusion denoiser, it was necessary to experimentally determine the lower bound of \(\epsilon\) and establish a suitable value accordingly.

To this end, we conducted experiments by varying \(\epsilon\) from 1 to \(10^{-7}\) in steps of \(10^{-1}\), as illustrated in Figure~\ref{epsilon}. Our observations revealed that at \(\epsilon = 10^{-6}\), both the PSNR and LPIPS metrics exhibited a sharp decline in performance, indicating the onset of numerical instability. This suggests an indirect lower bound for \(\epsilon\), specifically \(\epsilon > 10^{-6}\). For values above this lower bound, the diffusion denoiser demonstrated considerable robustness to variations in \(\epsilon\). Among these, we selected \(\epsilon = 10^{-3}\), which yielded the highest PSNR and LPIPS scores. To further mitigate numerical errors, we scaled this value by multiplying it with \(\max(x_{\text{noisy}})\), ensuring stability in practical applications.

\begin{figure}
    \centering
    \includegraphics[width=0.8\linewidth]{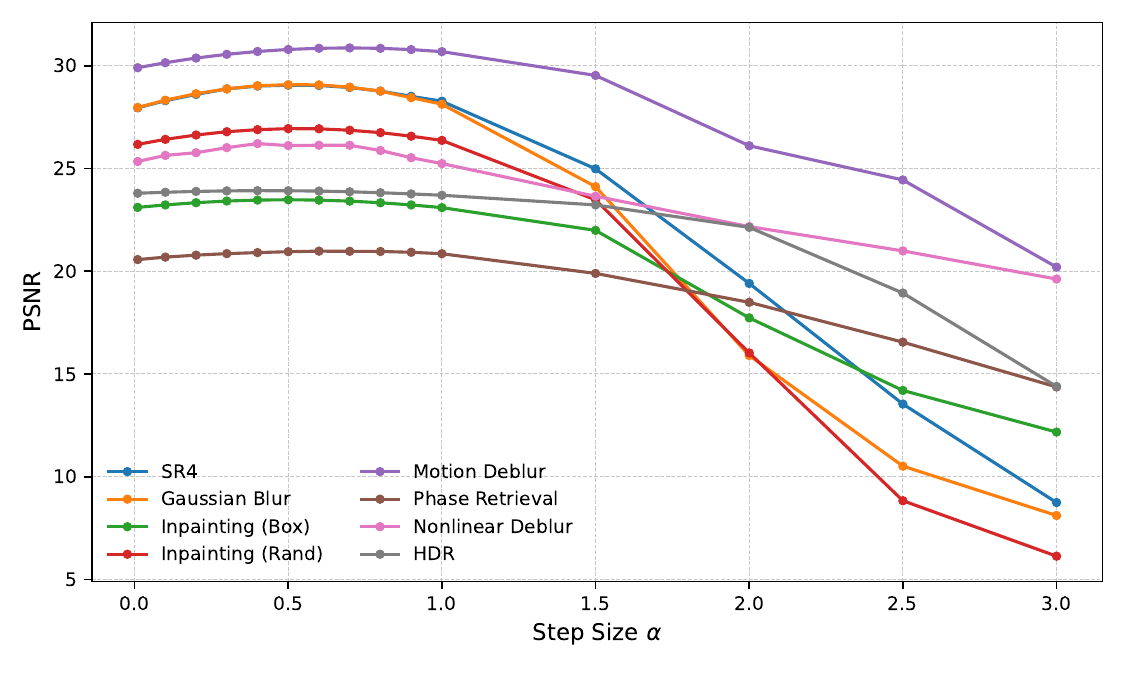}
    \caption{Sensitivity Analysis of the SURE Gradient Update Step Size ($\alpha$) on PSNR Performance for General Inverse Problem Tasks.}
    \label{alpha}
\end{figure}

\subsection{Hyperparameter \texorpdfstring{$\alpha$}{alpha}}
We performed a sensitivity analysis for the step size ($\alpha$) used in the SURE gradient updates. This analysis was conducted across general inverse problem tasks to identify an optimal and stable value for $\alpha$. For this analysis, we experimented by varying $\alpha$ across a range of values, as depicted in Figure~\ref{alpha}, and evaluated the impact on PSNR. Our findings consistently showed across all tasks that PSNR performance was stable and superior when $\alpha < 1$. Specifically, the optimal performance was typically observed in the range of $\alpha \approx [0.4, 0.6]$. In contrast, when $\alpha \geq 1$, there was a significant degradation in PSNR. This behavior indicates that an overly large $\alpha$ can lead to instability in the SURE gradient, adversely affecting convergence. Based on these empirical results, an effective upper bound for $\alpha$ is indicated as $\alpha < 1$. From the range of values that yielded robust performance, we selected $\alpha = 0.5$. This value consistently offered a favorable balance between high PSNR scores and convergence stability.

\section{Validation of PCA-based Noise Level Estimation}
\label{sec:Validation of PCA-based Noise Level Estimation}

\begin{figure}
    \centering
    \includegraphics[width=0.5\linewidth]{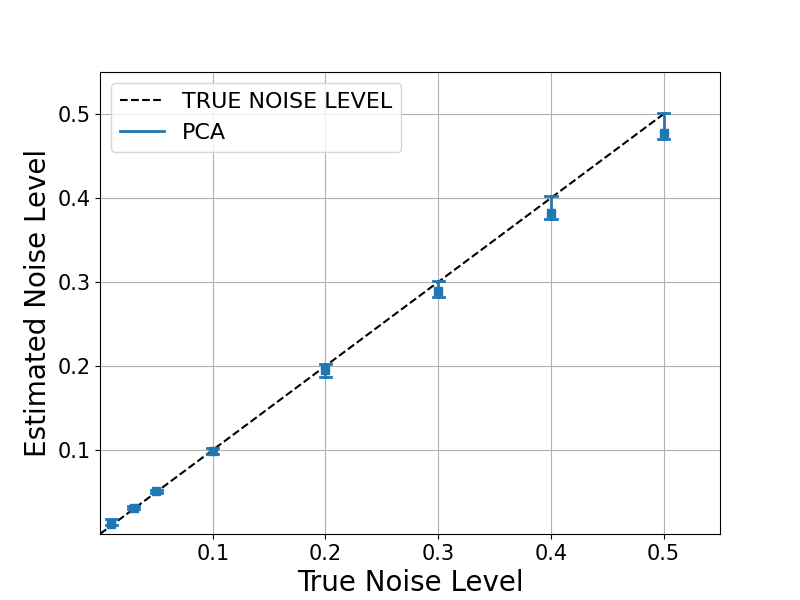}
    \caption{Comparison of true noise levels and PCA-based estimated noise levels across various noise levels, averaged over 100 samples per noise level. The dashed line represents the true noise level, and the blue markers with error bars show the PCA-estimated noise level, where markers indicate the mean and error bars represent the min-max range.}
    \label{PCA}
\end{figure}

We evaluated the PCA-based noise level estimation method on a synthetic dataset constructed by adding various Gaussian noise levels, ranging from \( \sigma = 0.01 \) to \( \sigma = 0.5 \) (\( \sigma = 0.01, 0.03, \ldots, 0.5 \)), to 100 FFHQ samples. These noise levels span eight distinct values, covering the range assumed to be present in samples reconstructed by the diffusion denoiser. As shown in Figure~\ref{PCA}, the estimated noise levels closely followed the true noise levels, as indicated by the near-diagonal trend in the plot, demonstrating excellent performance in predicting the true noise level across various noise levels and confirming the method's effectiveness in estimating the amount of noise present in the observation.

\begin{figure}
\centering
\includegraphics[width=\textwidth]{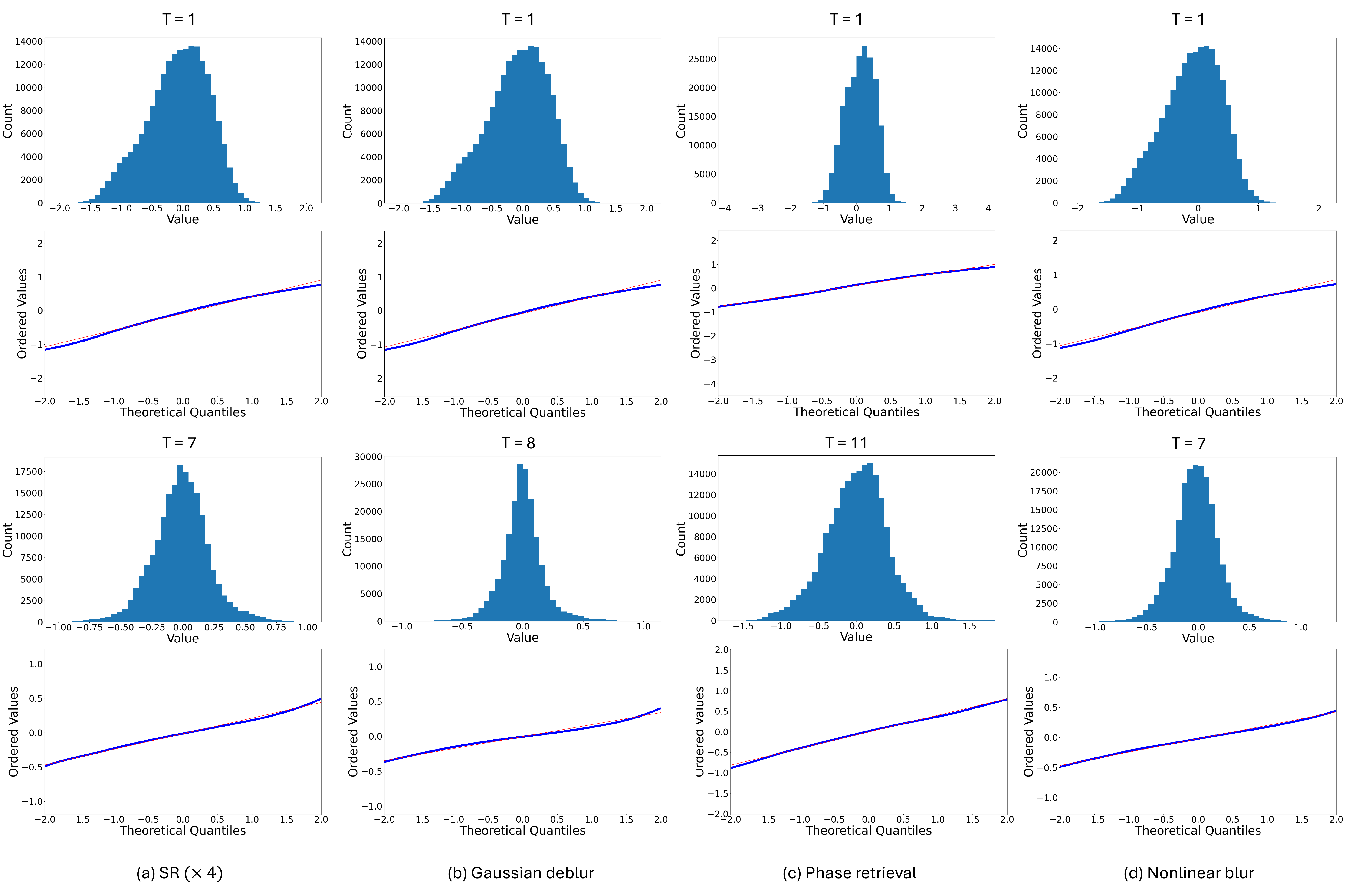}
\caption{\textbf{Histograms and Quantile-Quantile plots of residuals between $\hat{\mathbf{x}}_{0|t}$ and ground truth images across selected inverse problems at various sampling steps} (from a total of 16 steps). Experiments were conducted without the SURE gradient update in SGPS.}
\label{histogram}
\end{figure}
\section{Residual Noise Characteristics}

To empirically validate the Gaussian nature of the residual noise between the ground truth image $\mathbf{x}_0$ and the denoiser’s estimate $\hat{\mathbf{x}}_{0|t}$, we performed an analysis without the SURE gradient update in SGPS. This approach isolates the intrinsic residual characteristics produced by the diffusion model denoiser, independent of active corrections from SURE. Figure~\ref{histogram} illustrates histograms and Quantile-Quantile (Q-Q) plots of these residuals ($\mathbf{x}_0 - \hat{\mathbf{x}}_{0|t}$) for two linear inverse problems (SR($\times 4$) and Gaussian deblur) and two nonlinear inverse problems (Phase retrieval and Nonlinear deblur). Each task is examined at an early sampling step (e.g., T=1) and a later sampling step (e.g., T=7 or T=11, out of a total of 16 steps).

\begin{figure}
\centering
\includegraphics[width=\textwidth]{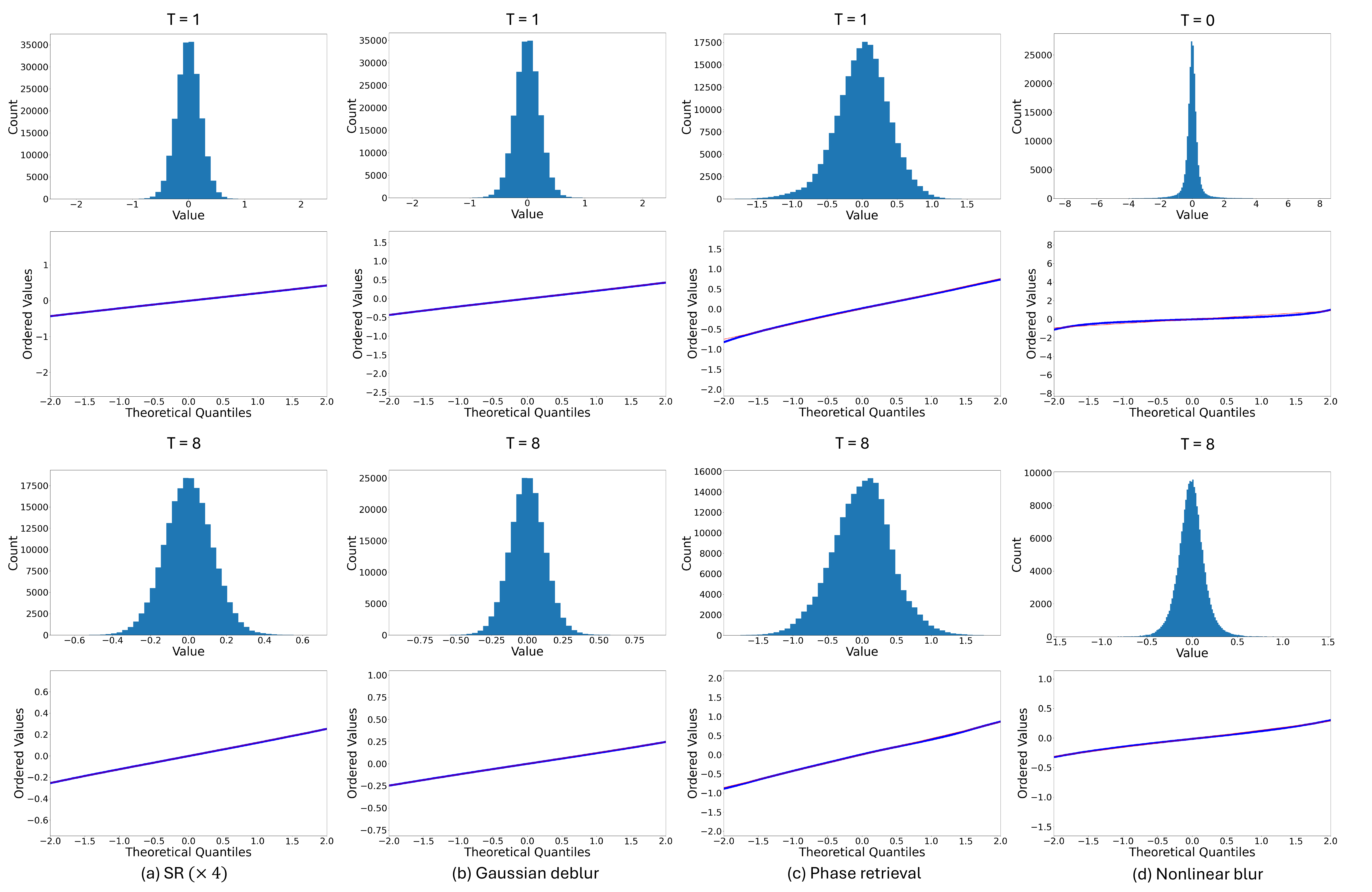}
\caption{\textbf{Histograms and Quantile-Quantile plots of residuals between $\hat{\mathbf{x}}_{0|t,y}$ and ground truth images across selected inverse problems at various sampling steps} (from a total of 16 steps). Experiments were conducted without the SURE gradient update in SGPS.}
\label{histogram2}
\end{figure}

Figure~\ref{histogram} clearly indicates that residual distributions across all tested inverse problems and sampling steps are bell-shaped, centered around zero, and closely resemble a Gaussian distribution. The accompanying Q-Q plots quantitatively reinforce this observation. Ideally, residuals from a perfect Gaussian distribution align precisely along the red diagonal line representing theoretical quantiles of a standard normal distribution. The empirical residuals (blue points) consistently follow this red diagonal line, primarily within the central distribution regions. Tail deviations occur, reflecting typical characteristics of empirical noise distributions, but do not significantly undermine the overall strong Gaussian approximation. Importantly, the Gaussian-like behavior remains stable from higher noise levels at initial steps to moderate noise levels in later steps.

This empirical validation of the residual noise’s Gaussian nature is critical as it forms the theoretical foundation for applying methods such as SURE, which inherently rely on Gaussian noise assumptions to effectively guide and correct the sampling process. The robustness of this near-Gaussian approximation, consistent across various tasks, underscores its practical applicability in diffusion-based inverse problem-solving.

Further analysis was conducted to examine whether the residual noise introduced during conditional guidance, defined as $\mathbf{z} = \hat{\mathbf{x}}_{0|t,y} - \mathbf{x}_0$, follows a Gaussian distribution. Figure~\ref{histogram2} provides histograms and Q-Q plots of these residuals for the same linear inverse problems (SR($\times 4$) and Gaussian deblur) and nonlinear inverse problems (Phase retrieval and Nonlinear deblur) at early and later sampling steps. The histograms again show bell-shaped distributions centered near zero, suggesting a Gaussian-like nature. This visual assessment is quantitatively validated through Q-Q plots, where empirical residuals (blue points) generally align with the theoretical quantiles of a standard normal distribution (red diagonal line), particularly within the central region. This near-Gaussian behavior consistently appears at both early and later sampling steps. Such empirical evidence substantiates the Gaussian noise assumption even after conditional guidance, crucial for the effective application of SURE gradient updates, thereby supporting its role in enhancing posterior sampling accuracy.

\section{Additional Experiments}
We conduct four controlled experiments to validate key design choices, using the representative $\times4$ super-resolution setting with 20 diffusion steps unless otherwise stated.

\begin{table}[h]
\centering
\caption{Comparison of one-step estimation vs. multi-step ODE solvers.}
\label{tab:reverse_diffusion_nfe}
\begin{tabular}{l c c c c}
\toprule
Reverse Diffusion & NFE per Step & Total NFE & PSNR $\uparrow$ & LPIPS $\downarrow$ \\
\midrule
One-step Estimation        & 3 (1 + 2 for SURE)    & 60   & \textbf{29.106} & \textbf{0.192} \\
3-step Euler ODE Solver    & 5 (3 + 2 for SURE)    & 100  & 28.922          & 0.199          \\
5-step Euler ODE Solver    & 7 (5 + 2 for SURE)    & 140  & 28.628          & 0.202          \\
\bottomrule
\end{tabular}
\label{tab:multi_ode}
\end{table}

\paragraph{One-step vs.\ multi-step ODE integration.}
We compare our one-step estimation with multi-step Euler ODE solvers by varying the number of integration steps while keeping other components unchanged (Table~\ref{tab:multi_ode}).
The one-step estimation achieves the best trade-off, yielding the strongest PSNR/LPIPS with the lowest NFE.
Increasing the number of Euler steps does not improve performance and can slightly degrade it, suggesting that additional discretized integration provides limited benefit in our setting.

\begin{table}[ht]
\centering
\caption{Comparison of using single vs.\ multiple vectors for Jacobian trace estimation.}
\label{tab:jacobian_trace_vectors}
\begin{tabular}{c c c c c}
\toprule
Number of Vectors & NFE per Step & Total NFE & PSNR $\uparrow$ & LPIPS $\downarrow$ \\
\midrule
1 & 3 (1 + 2) & 60  & \textbf{29.206} & \textbf{0.192} \\
3 & 5 (1 + 4) & 100 & 29.201          & 0.192          \\
5 & 7 (1 + 6) & 140 & 29.205          & 0.192          \\
\bottomrule
\end{tabular}
\label{tab:multi_vec}
\end{table}

\paragraph{Number of random vectors for trace estimation.}
We vary the number of random vectors used for Jacobian-trace estimation (Table~\ref{tab:multi_vec}).
Using more vectors does not lead to meaningful gains, while the NFE increases linearly with the number of vectors.
Therefore, we use a single random vector as the default choice.

\begin{table}[h]
\centering
\caption{Comparison of single vs.\ iterative SURE gradient updates.}
\label{tab:sure_updates}
\begin{tabular}{c c c c c}
\toprule
Number of SURE Updates & NFE per Step & Total NFE & PSNR $\uparrow$ & LPIPS $\downarrow$ \\
\midrule
1 & 3 (1 + 2)     & 60  & \textbf{29.206} & \textbf{0.192} \\
3 & 7 (1 + 3$\times$2) & 140 & 28.196          & 0.224          \\
5 & 11 (1 + 5$\times$2) & 220 & 28.199         & 0.224          \\
\bottomrule
\end{tabular}
\label{tab:multi_grad}
\end{table}

\paragraph{Iterative SURE gradient updates.}
We study whether performing multiple SURE-based correction steps improves reconstruction.
We run iterative updates using SGD (learning rate 0.5), and re-estimate the noise level via our PCA-based estimator at each iteration.
As shown in Table~\ref{tab:multi_grad}, iterative updates do not improve accuracy and tend to slightly degrade it, while substantially increasing NFE.
This supports using a single correction step.

\begin{table}[h]
\centering
\caption{Sensitivity analysis on the noise-level estimator $\hat{\sigma}_0$.}
\label{tab:sigma0_scaling}
\begin{tabular}{l c c}
\toprule
Scaling Factor for $\hat{\sigma}_0$ & PSNR $\uparrow$ & LPIPS $\downarrow$ \\
\midrule
0.5 (Underestimation) & 28.360 & 0.219 \\
0.8 (Underestimation) & 28.901 & 0.197 \\
\textbf{1.0 (Proposed)} & \textbf{29.206} & \textbf{0.192} \\
1.2 (Overestimation)  & 28.894 & 0.206 \\
1.5 (Overestimation)  & 23.496 & 0.414 \\
\bottomrule
\end{tabular}
\label{tab:pca_sensitivity}
\end{table}

\paragraph{Sensitivity to noise-level estimation.}
We evaluate the sensitivity of our method to noise-level estimation errors by multiplying the estimated $\hat{\sigma}_0$ by a constant factor (Table~\ref{tab:pca_sensitivity}).
Our method is robust to moderate misestimation: the performance changes only slightly under $\times0.8$ and $\times1.2$, indicating that precise estimation of $\hat{\sigma}_0$ is not critical for strong results.
As expected, larger errors—especially substantial overestimation (e.g., $\times1.5$)—cause a more noticeable performance drop.

\clearpage

\begin{figure}
    \centering
    \includegraphics[width=\textwidth]{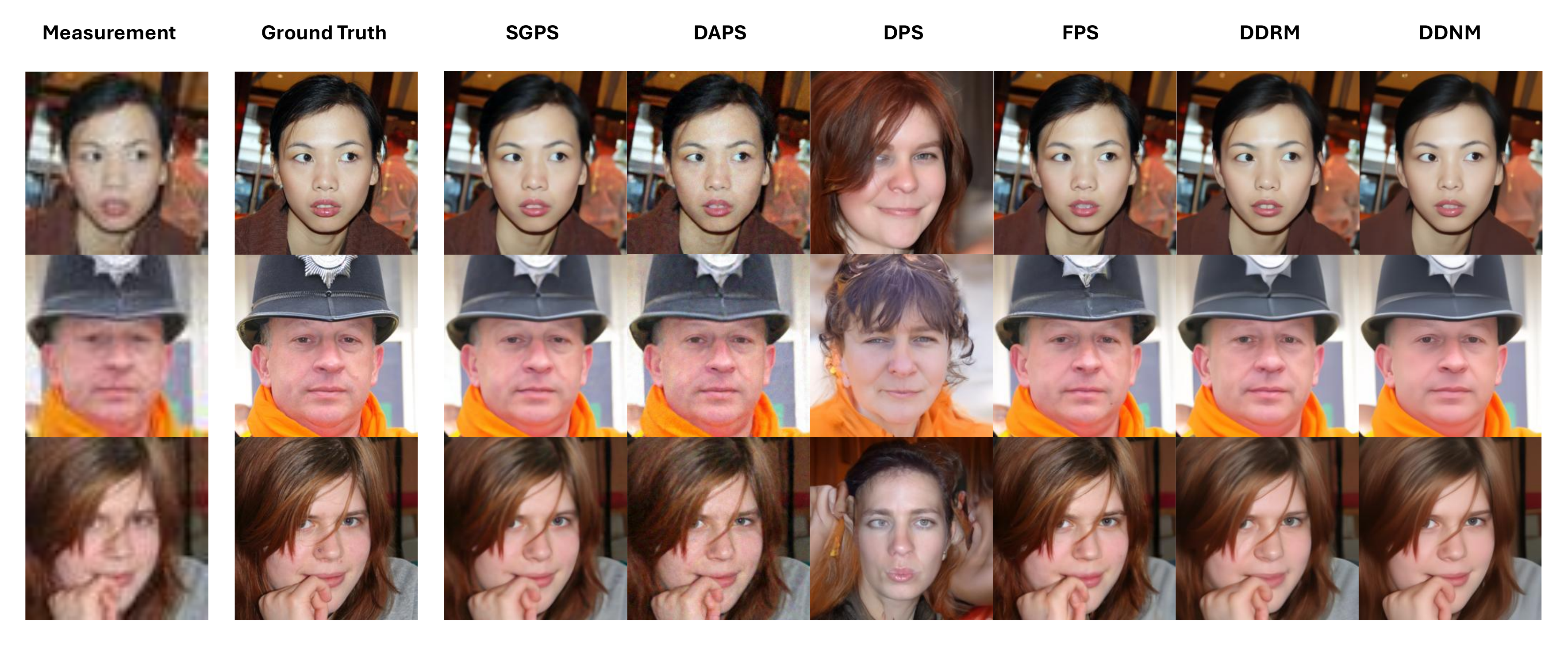}
    \vspace{-0.8cm} % Keep reduced space before next figure
    \label{SR4}
\end{figure}
\begin{figure}
    \centering
    \includegraphics[width=\textwidth]{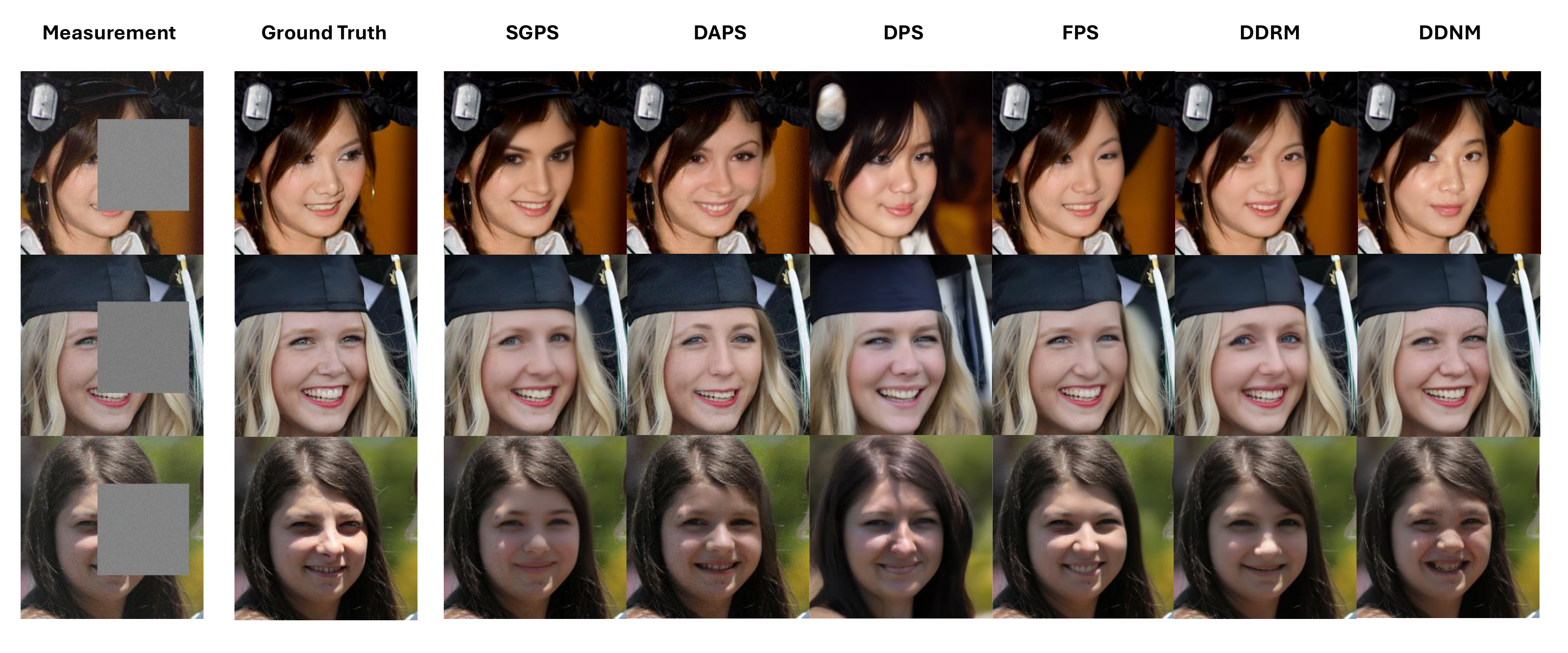}
    \vspace{-0.8cm} % Keep reduced space before next figure
    \label{inp_box}
\end{figure}
\begin{figure}
    \centering
    \includegraphics[width=\textwidth]{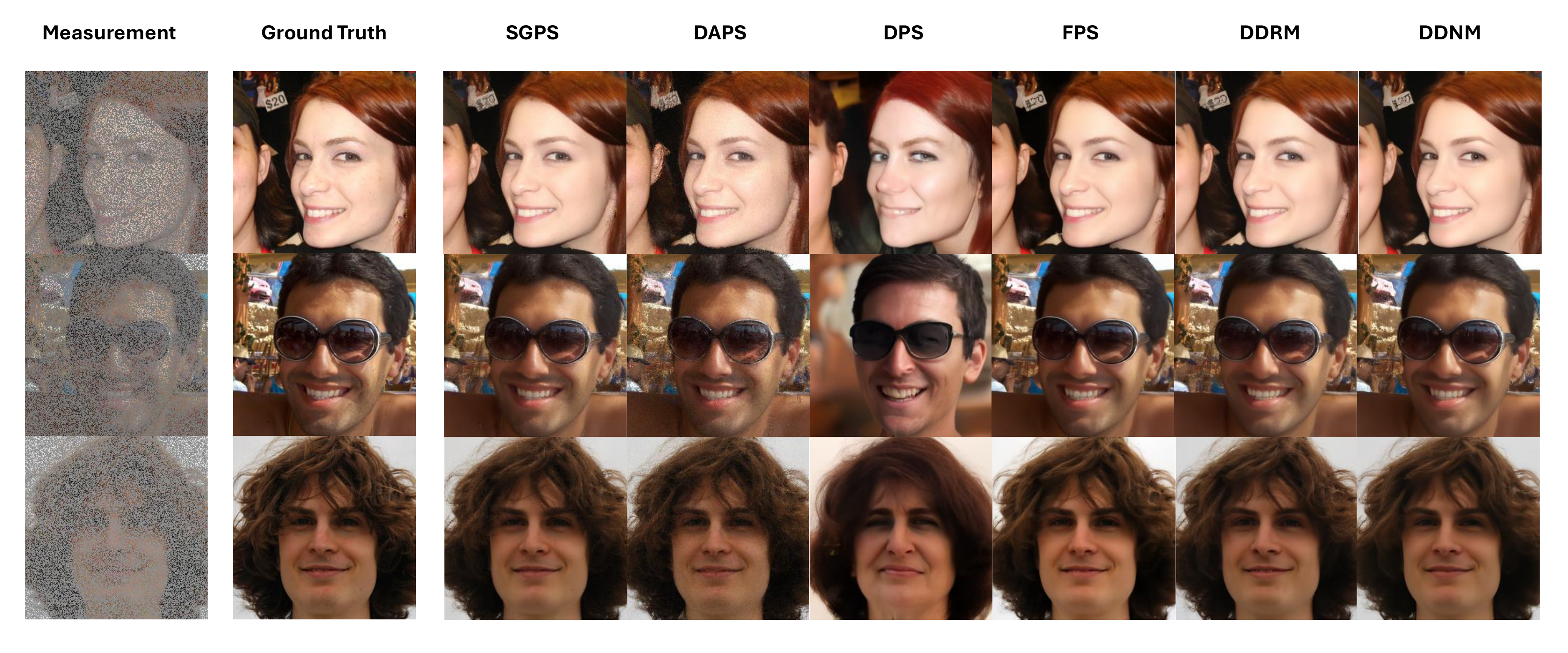}
    \vspace{-0.8cm} % Reduce space between image and its caption
    \caption{Sample comparison of all baselines including SGPS on the FFHQ 256 $\times$ 256 dataset under NFE100 conditions. From top to bottom: SR4, box inpainting, and random inpainting tasks.}
    \label{inp_rand}
\end{figure}
\clearpage

\end{document}